\documentclass[journal]{IEEEtran}
\usepackage{color}

\ifCLASSINFOpdf
  \usepackage[pdftex]{graphicx}
  \graphicspath{{../pdf/}{../jpeg/}}
  \DeclareGraphicsExtensions{.pdf,.jpeg,.png}
\fi
\usepackage{float}
\usepackage{makecell}
\usepackage{longtable}
\usepackage{stfloats}
\usepackage{multirow}
\usepackage{amsmath}
\usepackage{graphicx}
\usepackage{bm}
\usepackage{subfigure}
\usepackage[numbers,sort&compress]{natbib}

\usepackage{balance}
\usepackage{stfloats}
\usepackage[colorlinks,citecolor=blue,urlcolor=blue,bookmarks=false,hypertexnames=true]{hyperref} 

\begin{document}
%\small

\title{RaLiBEV: Radar and LiDAR BEV Fusion Learning for Anchor Box Free Object Detection Systems}

%\title{RaLiBEV: Radar and LiDAR BEV Fusion for Anchor Box Free Object Detection}

%\title{RaLiBEV2: On the Radar and LiDAR BEV Fusion base Object Detection}

\author{\IEEEauthorblockA{
Yanlong Yang\IEEEauthorrefmark{1},
Jianan Liu\IEEEauthorrefmark{1}\IEEEauthorrefmark{2},
Tao Huang\IEEEauthorrefmark{2},~\IEEEmembership{Senior Member,~IEEE,}\\
Qing-Long Han,~\IEEEmembership{Fellow,~IEEE},
Gang Ma, and 
Bing Zhu,~\IEEEmembership{Member,~IEEE}
}
\vspace{-5 mm}

\thanks{This work has been submitted to the IEEE for possible publication. Copyright may be transferred without notice, after which this version may no longer be accessible.}
\thanks{\IEEEauthorrefmark{1}Both authors contribute equally to the work and are co-first authors.}
\thanks{\IEEEauthorrefmark{2}Corresponding authors.}
%\thanks{The work of Y.~Yang and G.~Ma was supported by Beijing Municipal Science and Technology Commission under grant Z201100003920003.}
\thanks{Y.~Yang and G.~Ma are with Vanjee Technology, Beijing, 100193, P.R.~China. Email: \{yangyanlong, magang\}@wanji.net.cn.}
\thanks{J.~Liu is with Vitalent Consulting, Gothenburg, 45197, Sweden. Email: jianan.liu@vitalent.se.}
\thanks{T.~Huang is with the College of Science and Engineering, James Cook University, Cairns, QLD 4870, Australia. Email: tao.huang1@jcu.edu.au.}
\thanks{Q.-L.~Han is with the School of Science, Computing and Engineering Technologies, Swinburne University of Technology, Melbourne, VIC 3122, Australia. Email: qhan@swin.edu.au.}
\thanks{B.~Zhu is with the School of Automation Science and Electrical Engineering, Beihang University, Beijing, 100191, P.R.~China. Email:
zhubing@buaa.edu.cn.}
}

%\markboth{IEEE Robotics and Automation Letters, Submitted in May~2023}%
%\markboth{IEEE Transactions on Intelligent Vehicles, Submitted Dec~2023}
\markboth{  }
%\markboth{IEEE Transactions on Industrial Informatics, Submitted Nov~2022}
%\markboth{IEEE Transactions on Neural Networks and Learning Systems, Revised Oct.~2023}
{\MakeLowercase{\textit{et al.}}: Demo of IEEEtran.cls for IEEE Journals}

\maketitle

\begin{abstract}

In autonomous driving, LiDAR and radar are crucial for environmental perception. LiDAR offers precise 3D spatial sensing information but struggles in adverse weather like fog. Conversely, radar signals can penetrate rain or mist due to their specific wavelength but are prone to noise disturbances. Recent state-of-the-art works reveal that the fusion of radar and LiDAR can lead to robust detection in adverse weather. The existing works adopt convolutional neural network architecture to extract features from each sensor data, then align and aggregate the two branch features to predict object detection results. However, these methods have low accuracy of predicted bounding boxes due to a simple design of label assignment and fusion strategies. In this paper, we propose a bird's-eye view fusion learning-based anchor box-free object detection system, which fuses the feature derived from the radar range-azimuth heatmap and the LiDAR point cloud to estimate possible objects. Different label assignment strategies have been designed to facilitate the consistency between the classification of foreground or background anchor points and the corresponding bounding box regressions. Furthermore, the performance of the proposed object detector is further enhanced by employing a novel interactive transformer module. The superior performance of the methods proposed in this paper has been demonstrated using the recently published Oxford Radar RobotCar dataset. Our system's average precision significantly outperforms the state-of-the-art method by 13.1\% and 19.0\% at Intersection of Union (IoU) of 0.8 under 'Clear+Foggy' training conditions for 'Clear' and 'Foggy' testing, respectively.

\end{abstract}


\IEEEpeerreviewmaketitle

\section{Introduction}

\begin{figure}[t]
    \centering
    \includegraphics[scale=0.41]{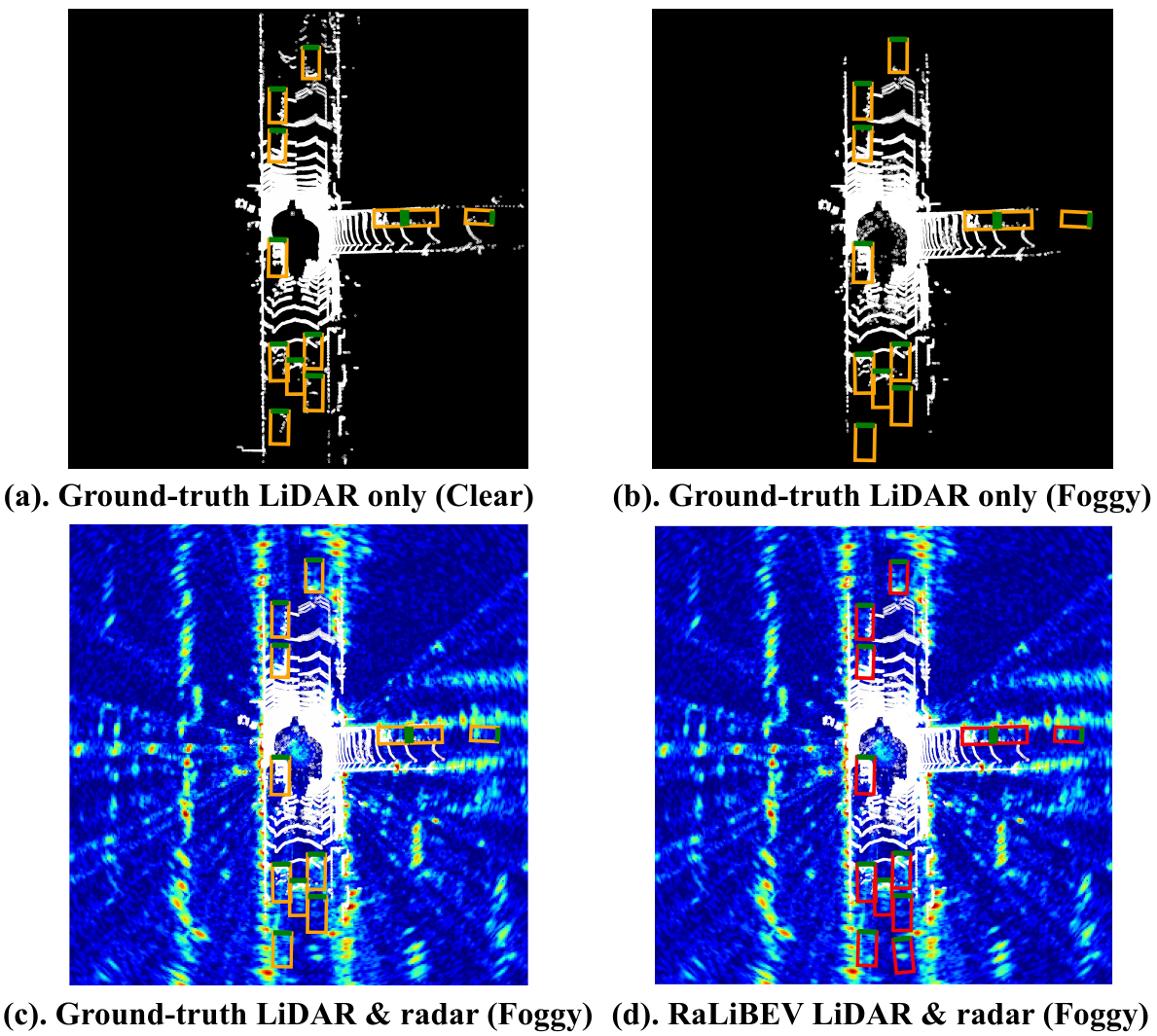}
    \caption{Performance of the RaLiBEV in clear and foggy weather. The LiDAR and radar visualization results are combined by radar range azimuth heatmap with jet pseudo-color in the background and LiDAR with white point cloud and 2D object bounding boxes. The ground-truth boxes are orange, and the predicted boxes by RaLiBEV are red. All the boxes use green lines to indicate the heading direction.}
    \label{RaLiBEV_Outcome_Visualization}
\end{figure}

\IEEEPARstart{A}{utonomous} driving systems typically employ sensors such as LiDAR, cameras, and radar to construct their perception functionality, which serves as the essential for the downstream tasks like trajectory prediction and ego motion planning \cite{CAA_Traj_Pred,CAA_Ego_Motion_Planning}, etc. Such perception scheme has been widely demonstrated in numerous works, covering various aspects such as data collection/generation \cite{Oxford_Robotcar_Dataset,scenario_generation_AOLi}, algorithm development \cite{LiDAR_based_ego_localization_TIV}, and real-world testing \cite{real_world_testing}.
LiDAR, a key sensor used in Advanced Driver Assistance Systems (ADAS) and autonomous driving, offers rich semantic and geometric information for environmental perception tasks \cite{PointCloudSurvey}. %,PointCloudSurvey_TNNSL}. 
These tasks include object detection \cite{LiDAR_object_detection_PointPillar,LiDAR_object_detection_CenterPoint}, multi-object tracking \cite{LiDAR_based_MOT_1,LiDAR_based_point_target_MOT_LEGO}, ego localization \cite{LiDAR_based_ego_localization_TIV,LiDAR_based_ego_localization_2}, and mapping \cite{LiDAR_based_SLAM_1}.
Recent works show that fusing LiDAR and camera data at the raw stream level can significantly improve the robustness of autonomous driving perception systems  \cite{LiDAR_and_camera_fusion_perception_3_TIV,LiDAR_and_camera_fusion_perception_4_TIV}. 
However, the performance of either camera or LiDAR can severely drop in adverse weather, i.e., rain, snow, and fog \cite{camera_object_detection_rain_TIV,LiDAR_Adverse_Weather_Survey}. 
For instance, cameras can become blurred, the operating range of LiDAR might be significantly reduced, and the number of noise points might increase due to the impact of heavy snow or fog. As illustrated in Fig. \ref{RaLiBEV_Outcome_Visualization}(b), the number of noisy points around the ego vehicle increases while the far end LiDAR points disappear, compared to the LiDAR point cloud measured in clear weather, as shown in Fig. \ref{RaLiBEV_Outcome_Visualization}(a).

In contrast with cameras and LiDAR, radar can naturally avoid the influence of adverse weather since the millimeter wavelength signals can penetrate the rain, snow, and fog. Such characteristic makes radar a powerful complementary sensor in addition to either camera or LiDAR in ADAS and autonomous driving \cite{automotive_radar_survey}, and even in future cooperative perception \cite{cooperative_perception_survey}
. The typical Frequency Modulated Continuous Wave (FMCW) radar, widely used in automotive systems, can accurately separate objects in range and speed dimension but has a low angular resolution due to the limited number of antennas in the compact space. Thus, radar can only provide many sparser detection points with ambiguity in angular information compared to the dense LiDAR point cloud. Such a drawback caused limitations on its performance when applied in the perception system \cite{SMURF,LXL,4d_imaging_radar_camera_3d_object_detection_rcfusion,4D_radar_3D_MOT}
. 

To take full advantage of the robustness of radar in adverse weather, radar range-azimuth heatmap rather than detection points has been employed to fuse with LiDAR point cloud for vehicle detection in the recent proposed anchor box-based methods \cite{radar_LiDAR_fusion_object_detection_MVDNet,radar_LiDAR_fusion_object_detection_ST-MVDNet}. However, their methods suffer from generalization issues since their approaches require predefined anchor boxes and they need to be carefully chosen according to the dataset. Moreover, these methods have been shown to have unsatisfactory performance, especially in the performance with high accuracy requirements. 

To address these issues, we propose radar and LiDAR in Bird's Eye View (BEV) fusion model, dubbed as RaLiBEV, for object detection. RaLiBEV introduces a novel object detection approach that fuses radar and LiDAR data in BEV in anchor boxes free scheme, intended for real-time application. The LiDAR point cloud is transformed into a structured data format in BEV, a process known as pillarization, while the radar range-azimuth heatmap naturally provides discrete sampling in BEV. These BEV feature maps from both radar and LiDAR are then separately extracted and combined to create an aggregated BEV feature map. This map is processed through the network's backbone and then fed into a multi-scale detection head that doesn't rely on anchor boxes to estimate the bounding boxes of vehicles.

Unlike the anchor box-based approach, the anchor box-free detection head first identifies representative pixels, or key points, in the predicted feature map. These key points are then used as positive samples to regress the remaining feature channels. However, this approach can lead to inconsistencies between the classification of foreground and background, and the regression problem. Specifically, a key point with a high classification score doesn't necessarily result in better accuracy in bounding box regression. Another challenge lies in effectively fusing the rich feature information from the radar range-azimuth heatmap with the LiDAR feature map in a complementary manner, rather than simply concatenating them.

To address these issues, RaLiBEV incorporates innovative label assignment strategies to reduce the inconsistency between foreground-background classification and regression. Additionally, it employs interactive transformer-based BEV fusion techniques to better explore the interaction between radar and LiDAR feature maps. The experiment results demonstrate that our method outperforms the state-of-the-art method by 13.1\% and 19.0\% at IoU equals 0.8 in clear and foggy weather testing respectively.

The main contributions of this paper are concluded as below:
\begin{itemize}

\item A radar and LiDAR BEV fusion-based anchor box free object detector, RaLiBEV, is proposed. The RaliBEV can generate accurate 2D bounding boxes in BEV by fusion of features extracted from the radar range-azimuth heatmap and the LiDAR point cloud.

\item A novel label assignment strategy, Gaussian Area-based Consistent Heatmap and IoU Cost Positive Sample Assignment (GACHIPS), is proposed to combat the inconsistency between the classification of key points and the corresponding bounding box regression tasks in the anchor box free object detector.
% introduced dense query到里面，起到了拟合能力更好的效果
\item A novel interactive transformer-based fusion module, Dense Query Map-based Interactive BEV Fusion (DQMITBF), is proposed to fuse radar and LiDAR feature maps. With the introduction of dense, learnable queries and a symmetric fusion architecture, the fusion module provides an information-sensitive fusion mechanism to aggregate suitable features from both maps under various weather conditions.
%Several interactive transformer-based improved fusion modules are proposed to properly fuse radar and LiDAR feature maps by interacting with each other, resulting in significant performance improvements, particularly on precise object bounding box estimation. The Dense Query Map-based Interactive BEV fusion (DQMITBF) owns the best performance.
%
% \textcolor{red}{Tao: Do we have a corresponding study to show which one is the best and how much performance improvement compare to the baseline?}

\item The aforementioned object detector can achieve fairly accurate detection performance even in adverse weather, as shown in of Fig. \ref{RaLiBEV_Outcome_Visualization}(d) compared to the ground-truth bounding boxes illustrated in Fig. \ref{RaLiBEV_Outcome_Visualization}(c) which outperforms the state-of-the-art radar and LiDAR fusion-based object detector with a large margin in the popular Oxford Radar RobotCar (ORR) dataset \cite{Oxford_Robotcar_Dataset}.

\end{itemize}

The rest of this paper is organized as follows. Section \ref{sec related} discusses the related works, including LiDAR-based perception methods, radar-based perception methods, and the methods which fuse radar and LiDAR for object detection. In Section \ref{sec methods}, the proposed radar and LiDAR fusion-based object detection approach, RaLiBEV, is introduced. In addition, the proposed label assignment strategies and interactive transformer-based fusion schemes are also described. Then, the experiment process and corresponding results on ORR datasets are explained and analyzed in Section \ref{sec results}. Finally, the conclusion and further work are drawn in Section \ref{sec conclusion}.

% related works写法提纲：相关的工作有什么，怎么做，优缺点是什么
\section{Related Works}\label{sec related}

\subsection{LiDAR-based Object Detection}
% 1. Lidar is a wildly used sensor in environment perception, one of the most important mission in which is object detection. (Lidar是用来感知环境的传感器，而环境感知中的一个重要任务就是目标检测。)
LiDAR is a widely used sensor in surrounding environment perception, and object detection is one of the most important missions in environment perception. 
% 2. 基于深度学习的lidar目标检测方法按照input data处理的方式分为四类：point based method, grid based method, graph based method, combination based method
The LiDAR object detection method based on deep learning can be divided into four categories according to the processing approach on the input point cloud. 

% 3. point based method 使用线性层直接对乱序的点云进行特征提取。
The first approach is the point-based method. It implements a linear layer to extract features on unordered point cloud \cite{PointNet,PointNet++}. 
% 4. grid based method 使用了2D或者3D的grid将点云进行discretize，连接mature CNN进行特征数据的特征提取。
The second approach is the grid-based method. This method discretizes the point cloud into 2D grids or 3D voxels, then applies CNN for further feature extraction \cite{LiDAR_object_detection_VoxNet,LiDAR_object_detection_PointPillar,LiDAR_object_detection_CenterPoint,LiDAR_object_detection_VoxelSetTransformer}.
% 5. graph based method 将点云数据视作graph, 点作为vertics, relation between points as edge, applies GNN for feature extraction.
The third approach is the graph-based method, which transforms point cloud as graph data format \cite{PointGNN}. The Graph Neural Network (GNN) is employed for extracting the feature and detecting the objects, where the points are defined as vertices and the relations between points are defined as edges.
% 6. Combination method包括6.1：Discretize point cloud to 2d or 3d grid, then applies point based feature extraction in each grid; 6.2 Combine variant view such as range view and bird eye view for feature extraction and concatenation.
The fourth type is the combination method. One way is to combine features extracted from two different representations of the point cloud, such as points features and grid/voxel features \cite{LiDAR_object_detection_PDV}. Another way is to concatenate features from different views of 2D grids, such as range view and BEV, and decode concatenated features by 2D CNN afterward \cite{LiDAR_object_detection_MVFuseNet}. 
Although LiDAR can provide precise and dense 3D spatial information, the inherent problem of facing adverse weather still limits its performance.

\subsection{Radar-based Object Detection}
Radar, utilizing under-millimeter wave band electromagnetic signals, is capable of penetrating adverse weather conditions, making it effective for object detection under noise. Both conventional and deep learning methods are employed for this task.

Conventional methods, such as Constant False Alarm Rate (CFAR) algorithms \cite{CFAR} and clustering techniques \cite{DBSCAN}, dynamically adjust the detection threshold based on noise levels and group detected points into objects. However, these methods may struggle with complex environments and overlapping objects.
% segmentation+clustering
Deep learning methodologies initiate the processing of radar point clouds, employ semantic segmentation for categorization and utilize clustering techniques for the assembly of objects \cite{radar_instance_segmentation_1,radar_instance_segmentation_2}. This method provides detailed object detection but requires significant computational resources.
% pointnet point GNN
More recent approaches, such as PointNet models \cite{ordinary_radar_based_3d_bev_object_detection_point_processing_1,ordinary_radar_based_3d_bev_object_detection_point_processing_2} and Graph Neural Networks (GNNs) \cite{ordinary_radar_based_3d_bev_object_detection_graph_1,ordinary_radar_based_3d_bev_object_detection_graph_2}, process point cloud data directly. PointNet learns complex patterns for object detection and classification, while GNNs capture both local and global patterns for enhanced detection accuracy. These methods build a end-to-end points-based object detection pipeline for radar object detection. However both methods require substantial training data and computational resources.
% clustering and classification on heatmap or directly process heatmap
Lastly, range-azimuth heatmaps are used for radar object detection. One approach applies clustering and classification to the heatmap \cite{radar_object_detection_with_signal_DL_1,radar_object_detection_with_signal_DL_2}, while another uses deep learning models to directly predict object presence and location from the heatmap \cite{ordinary_radar_based_3d_bev_object_detection_heatmap_1,ordinary_radar_based_3d_bev_object_detection_heatmap_6,ordinary_radar_based_3d_bev_object_detection_heatmap_7,ordinary_radar_based_3d_bev_object_detection_heatmap_8,ordinary_radar_based_3d_bev_object_detection_heatmap_9}. Methods leveraging heatmap-based techniques possess the capability to extract the most comprehensive information, albeit at the cost of increased computational resources.

% end with the drawbacks of radar only object detection
Despite its ability to penetrate adverse weather, radar's limitations, such as susceptibility to multi-path propagation and lower resolution, hinder detailed object recognition. These drawbacks underscore the necessity of sensor fusion in autonomous driving systems. The fusion of radar and LiDAR data, leveraging LiDAR's high-resolution point cloud data, can significantly enhance object detection and recognition capabilities. 

\subsection{Radar and LiDAR Fusion-based Object Detection}

Radar and LiDAR both have their strength and weaknesses. Fusing the information from the two types of sensors can achieve a better surrounding perception \cite{DEF,radar_LiDAR_fusion_object_detection_MVDNet,radar_LiDAR_fusion_object_detection_ST-MVDNet}. Generally, sensor fusion can be performed at the result or feature levels.

In result-level fusion, each sensor is characterized as an individual detection union. Then the detected objects are collected to generate the final result. This method consumes less transmission bandwidth and has a fast operation speed. However, its performance is severely limited by each single sensor detector.

% In feature-level fusion, the deep learning-based method is often applied. The deep learning neural network extracts the features from the two sensor data. Then the extracted features are aligned and fused 

% 在radar与Lidar的融合的网络设计中，需要将来自两个传感器的特征信息聚合到一起，为端到端网络提供丰富而且可靠的信息
In feature-level fusion, it is essential to aggregate feature information from both sensors, providing rich and reliable data for end-to-end networks \cite{4d_imaging_radar_lidar_3d_object_detection_1,4d_imaging_radar_lidar_3d_object_detection_2}.
% 传统的融合策略将来自不同传感器的数据经过CNN网络进行特征提取之后，简单进行拼接或加和[BEVFusion]。这类融合设计计算速度较快，但是对于数据的特征信息的加权组合能力有限。
General fusion strategies, such as concatenating or adding data from different sensors after CNN-based feature extraction, offer faster computational speeds but limited capacity in terms of weighted combination of feature information.
% 而使用transformer中的attention机制进行信息的抓取information scrap已经被证明是具有更好有效性的设计。
Utilizing the attention mechanism proposed in transformers for information scraping has been demonstrated to be more effective \cite{self_attention}.
% 然而，先前的工作采用cross-attention无论是在不同数据通道上，或者是不同时序数据上[MVDNet系列]，都存在一定程度的有效性问题。因为单纯的attention不依靠query来抓取信息，限制了模型对于难样本的拟合能力。
However, previous works utilizing cross-attention, whether on different data channels or across different temporal data \cite{radar_LiDAR_fusion_object_detection_MVDNet,radar_LiDAR_fusion_object_detection_ST-MVDNet,radar_LiDAR_fusion_object_detection_ST-MVDNet++}, suffer from certain effectiveness issues. This is because relying solely on attention to capture information without query limits the model's flexibility to fit difficult samples.
% Bi-LRFusion提出采用双向的lidar与radar特征融合scheme，使用了query-based的方式，从高度视角和BEV视角引导lidar特征与radar特征融合。然而它采用的sparse query却严重限制了它在面对radar range-azimuth heatmap数据上的性能。1.稀疏 2.语义歧义 3.精准几何geometry and spatial information
Bi-LRFusion \cite{radar_LiDAR_fusion_object_detection_Bi-LRFusion} proposes a bidirectional LiDAR and radar feature fusion scheme, which uses a query-based approach guided by both elevation-view and bird's-eye-view perspectives for fusing LiDAR and radar features. However, its utilization of sparse queries severely restricts its performance when handling dense radar range-azimuth heatmap data. To tackle this problem, we propose a dense, learnable query map-based transformer block that merges features from each sensor. This approach offers dense and adaptable attention weights for enhanced sensor fusion.

% -------------------------------------------------------------------

\subsection{Label Assignment in Object Detection}
% label assignment问题的提出是为了解决在目标检测的输出头监督过程中，确定哪些anchor box或者anchor point可以作为正样本。通过计算这些正样本与真值的loss，从而实现模型的梯度计算，更新模型参数。
%\textcolor{red}{ljn: what is gt? what is GT? Which reference paper corresponds to which approach, e.g., ATSS, AutoAssign, PAA and OTA?}

% 传统的label assignment设计最初来自于anchor based和anchor free的目标检测框架。在这些框架中，他们通过一些intuitive的cost，例如anchor box与gt box的IoU，或者是anchor point是否在gt box内部，使用固定的阈值来确定其是否为正样本[fast-rcnn][yolo][FCOS][centernet][cornernet]etc.。
The traditional label assignment design was initially proposed in anchor-based and anchor-free object detection frameworks. In these frameworks, fixed thresholds were used to determine whether a box or point is a positive sample, based on intuitive costs such as the IoU between anchor boxes and ground-truth boxes or whether an anchor point is inside a gt box. 
% 然而，这些Heuristic（启发式）的设计存在的最大问题，就是忽略了分类和框回归一致性(consistency)的问题。既分类分数高的anchor box或者anchor point，框回归并不一定是最好。这将导致模型在推理过程中可能选取较差的框当作输出，降低模型的高精度检测性能。
However, the biggest problem with these heuristic designs is that they ignore the issue of consistency between foreground-background classification and box regression \cite{label_assignment_inconsistency_problem}, meaning that high-scoring anchor boxes or points may not always lead to the best box regression results. This gap can cause the model to select suboptimal boxes as output during inference, which lowers the model's detection performance.

% 在anchor-based的框架下，由于所有与gt框IoU超过一定阈值的anchor box都要计算loss，并回归。而它们的loss设计为分类与框回归相加的loss。这样的平行loss设计使得分类与框回归毫无关联，进而导致了上述的一致性问题。
In anchor-based frameworks \cite{label_assign_fast-rcnn,label_assign_yolo}, since all anchor boxes with an IoU greater than a certain threshold with the gt box need to calculate loss and regression, the loss is designed as a sum of foreground-background classification and box regression losses. The parallel loss design makes the two part independent of each other, leading to the aforementioned consistency problem.
% 同理，在anchor-free的框架下，固定阈值的label assignment采用使用某些存在于gt box内部的固定anchor point作为正样本，例如中心点[centernet]或者角点[cornernet]。在平行地分别对分类和框回归分别计算loss之后相加，二者也存在回归时候的数值gap。同样存在上述的不一致性问题。
Similarly, in anchor-free frameworks \cite{label_assign_FCOS,label_assign_centernet,label_assign_cornernet}, the fixed threshold label assignment uses some fixed anchor points inside the gt box as positive samples, such as center points or corner points. After calculating foreground-background classification loss and box regression loss separately and adding them together, there is still a numerical gap in regression. Again, the inconsistency problem prevails.

% 在发现这个问题之后，有一些工作尝试来解决这样的问题，例如[freeanchor][ATSS][autoassign][PAA][OTA]等。
To address this problem, some works have attempted to focus on matching the predicted results and ground-truth by combining label assignment with model prediction, such as freeAnchor \cite{label_assign_freeanchor}, ATSS \cite{label_assign_ATSS}, AutoAssign \cite{label_assign_autoassign}, PAA \cite{label_assign_PAA} and OTA \cite{label_assign_OTA}.
% 这些方法将正负样本分配问题的解决办法从匹配anchor与gt，发展成与模型预测结合，匹配预测结果与真值，最大程度缩小了上述的训练阶段与测试阶段预测框解码的一致性。
These methods evolved the solution for assigning positive and negative samples from matching anchors and gt boxes to aligning predicted outcomes with ground-truth. This approach maximally reduces the consistency gap for bounding box decoding mentioned earlier.

% 然而这些方法关注的是图像中的目标检测，重点解决目标存在各种大小比例和互相遮挡等问题，为单个目标分配了多个正样本。然而一旦引入多正样本，就会不可避免地存在多个正样本的竞争，在测试阶段同样出现不可弥合的一致性gap。
However, these methods primarily concentrate on image-based object detection, addressing challenges such as varying object sizes, aspect ratios, and mutual occlusion by assigning multiple positive samples to individual gt targets. Nevertheless, incorporating multiple positive samples inevitably results in competition among them during the training process, as they all contribute to the gradient. This exacerbates the consistency gap during the inference phase.
% -------------------------------------------------------------------

\section{Proposed Method}\label{sec methods}
%In our work, we proposed the RaLiBEV and demonstrated its performance in the ORR dataset. The experiment result shows that in both clear and foggy weather scenarios, RaLiBEV performs significantly better than the \textcolor{blue}{(two)} previous state-of-the-art methods, especially in high-precision detection results. 
In this section, the  framework of the proposed anchor box-free radar and LiDAR fusion-based object detector are first introduced. Then, two enhancements of RaLiBEV, one with label-assignment strategies and the other with interactive transformer fusion strategies, are described.

%\subsection{Anchor Box Free Object Detection with Fusion of Radar and LiDAR in BEV}\label{Description of the overall network}

\subsection{Anchor Box Free Object Detection}\label{Description of the overall network}

\begin{figure*}[t]
    \centering
    \includegraphics[scale=0.55]{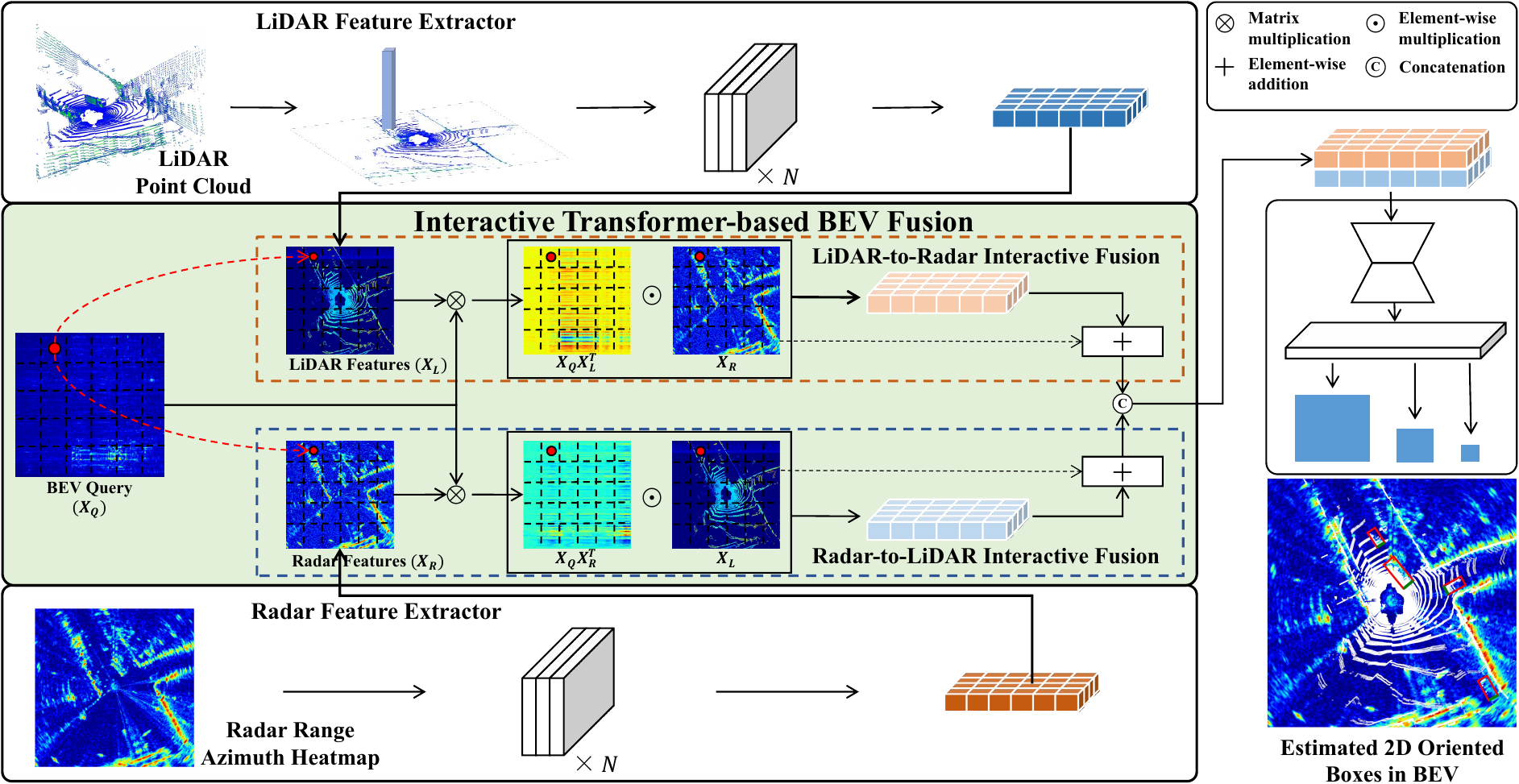}
    \caption{The entire pipeline of proposed radar and LiDAR fusion-based anchor box free object detector, RaLiBEV.}
    \label{RaLiBEV Framework}
\end{figure*}

Following the idea of sensor fusion, aggregating features from each sensor under BEV representation is straightforward. 
%\textcolor{red}{Therefore, based on the YOLOv4 \cite{yolov4} backbone and neck, an anchor box-free radar and LiDAR BEV object detector, RaLiBEV, is proposed, as shown in Fig. \ref{RaLiBEV Framework}. The corresponding LiDAR point cloud and radar range-azimuth heatmap are used as input data. Then the aggregated features are forwarded into the YOLOV4 backbone and neck for further processing. Finally, a multi-scale anchor box free detection head is employed to decode the 2D BEV rotating bounding boxes detection result.}
As depicted in Fig. \ref{RaLiBEV Framework}, the raw LiDAR point cloud and radar range azimuth heatmap undergo pre-processing via their respective sensor branch's feature extractors. These two features are then fed into fusion module for feature fusion. The combined features are subsequently processed by a YOLOV4-based \cite{yolov4} feature extractor and a multi-scale anchor-free object detection head to decode the 2D BEV rotating bounding boxes detection result. 

%For the radar data, the energy of the range-azimuth heatmap represents the possibility of object existence under the BEV space. Where the object exists, the energy of the heatmap would be greater than that of the empty area. 
Specifically, the radar data is converted from polar to Cartesian coordinates by applying bilinear interpolation, so the final structure shape of the data shows as $[W_r, H_r, C_r]$. The $W, H,C$ denote width, height and channel of the feature map respectively. For LiDAR data, since the point cloud is an unordered 3D points set, the data needs to be structured before subsequent operations. Therefore the method of PointPillars \cite{LiDAR_object_detection_PointPillar} is applied to form the input LiDAR data into a fixed shape $[W_l, H_l, C_l]$ under BEV.

After the pre-processing, $N$ convolution layers on each branch are applied to extract features without downsampling. Then two branch data are fused with an interactive transformer module and sent into the YOLOv4 backbone for further feature extraction.

A multi-scale anchor-free detection head is added at the end to decode the detection results. The first output channel of the detection head is a predicted heatmap, representing the object's existence probability. The second to the last channel represent $\triangle x$, $\triangle y$, width, length, angle classification, angle offset, and object class, respectively. $\triangle x$ and $\triangle y$ denote the predicted box center offset in x and y direction. Two channels are set to help angle regression. One is an angle classification channel, the other is an angle offset regression channel.
% 我需要在这个地方这么详细地描述检测头的通道设置吗

Following the state-of-the-art anchor-free detection paradigm, the positive anchor points of the feature map should be specified first, then start regression based on those locations. For each ground-truth bounding box in the BEV, a Gaussian distribution map is generated under each head scale. This serves as a supervision signal to train the network for heatmap prediction in a supervised learning approach. The generation method can be described by:
\begin{equation}
    \label{gaussian_map_generation_equation}
    \bm{G}(\bm{x}|\bm{\mu},\bm{\Sigma})=\frac{1}{2\pi^\frac{d}{2}\left|\bm{\Sigma}\right|^\frac{1}{2}}e^{-\frac{1}{2}(\bm{x}-\bm{\mu})^{T} \bm{\Sigma}^{-1} (\bm{x}-\bm{\mu})},
\end{equation}
where the $\bm{x}=[i,j]$ is the BEV spatial vector, and $i,j$ range from 0 to number of grid on the feature map. $d$ is the dimension of $\bm{x}$. $\bm{\mu}$ is the mean of $\bm{x}$, which is the center coordinates of ground-truth boxes. $\bm{\Sigma}$ is a covariance matrix of the distribution, which can be calculated by:
\begin{equation}
    \label{sigma_equation}
    \bm{\Sigma} = E(\bm{x}-E(\bm{x}))(\bm{x}-E(\bm{x}))^T,
\end{equation}
where the four corners' 2D coordinates of a box are used as $\bm{x}$, and the $E(\bm{x})$ equals the center coordinates of the box. Such operation is done iteratively over all ground-truth objects in one time frame, thus a ground-truth Gaussian mixture distribution is generated.

\subsection{Label Assignment Strategies}\label{label assignment strategy}
For an object detection problem, the total loss, in general, can be described as the sum of foreground-background classification loss and box regression loss, as shown here:
\begin{equation}
    \label{loss_formulation}
    \mathcal{L} = \mathcal{L}_{cls}+\mathcal{L}_{box},
\end{equation}
where the $\mathcal{L}_{cls}$ is the foreground-background classification loss implemented with focal loss \cite{focal_loss}, the $\mathcal{L}_{box}$ is the box regression loss implemented with smooth $L1$ loss. The focal loss is modified with Gaussian mixture distribution weight $G_{i,j}$:
\begin{equation}
%\footnotesize
    \label{focal_loss}
    \mathcal{L}_{cls} = \left\{
    \begin{aligned}
    &-\sum_{i,j=0}^M G_{i,j}(1-p_{i,j})^\gamma log(p_{i,j})~~~{\rm if}~G_{i,j}=1, \\
    &-\sum_{i,j=0}^M G_{i,j} p_{i,j}^\gamma log(1-p_{i,j})~~~~~~~\rm{otherwise}.
    \end{aligned}
    \right.
\end{equation}
$M$ is the grid number of feature map. $G_{i,j}\in[0, 1]$ specifies each element in the ground-truth Gaussian mixture distribution generated from Eq. (\ref{gaussian_map_generation_equation}). $p_{i,j} \in [0, 1]$ is the model's estimated value for each anchor point on BEV feature map, and $\gamma$ is a static parameter that balances the weight of positive and negative samples. By weighting the gradient with a Gaussian weight, the model's convergence will be more stable and quicker \cite{label_assign_centernet}.

The box loss is consist of $\triangle x$, $\triangle y$, width, length and angle offset $\triangle\theta$ loss, which are in the formulation of smooth $L1$ loss. While for the angle classification loss, the cross-entropy loss $l_{\theta}^{ce}$ is applied:
\begin{equation}
    \label{box_loss}
    \mathcal{L}_{box} = \sum_{i,j=0}^M\alpha_{i,j}(l_{\triangle xy}^{reg}+l_{wl}^{reg}+l_{\theta}^{ce}+l_{\triangle\theta}^{reg}).
\end{equation}
where $\alpha_{i,j} \in \{0,1\}$ is the binary element of mask matrix $\bm{\alpha}$. Typically, \boldsymbol{$\alpha$} is utilized to control which anchor points on the BEV feature map are selected for calculating the box loss. The chosen samples are referred as positive samples, denoting by $\alpha_{i,j} = 1$, while the unselected ones are termed as negative samples with $\alpha_{i,j} = 0$.
% 杨炎龙：由于只有正样本对最终的loss有贡献，确定哪些anchor points是positive sample变得尤为关键。这个问题就叫做label assignment问题。好的label assignment策略可以降低模型的漏检率，提高模型的检测高精度性能。而坏的label assignment策略则会导致模型在test时的NMS阶段选择回归更差的预测结果，这个问题称作分类与回归的不一致性问题。

The box regression loss could only occur on the positive samples, while the foreground-background classification loss is computed upon every anchor point on the feature map. This leads to the problem that an anchor point with a higher foreground-background classification score does not always index the best bounding box regression result during decoding the predicted object. This is known as the inconsistency problem in label assignments. A label assignment strategy with lesser inconsistency between foreground-background classification and box regression can decrease the model's miss-detection rate and help achieve higher detection accuracy.

%\textcolor{blue}{(After creating the ground-truth Gaussian map, the focal loss \cite{focal_loss} is implemented to regress the predicted heatmap. The label assignment strategies should be considered first to alleviate the inconsistency between the foreground-background classification and regression.)}

Typically, three strategies of label assignment could be applied, including bipartite matching, 
%
%which could be solved by the Hungarian algorithm (e.g., DETR \cite{DETR}), 
%
multi-positive label assignment, and single-positive label assignment. 
``Multi/single" indicates how many anchor points can be chosen as the positive sample for one object.

\subsubsection{Bipartite Matching Label Assignment}

The typical bipartite matching label assignment is based on the Hungarian algorithm. This set-based label assignment builds a cost matrix of IoU between predicted and ground-truth boxes, where the predicted boxes are decoded from a predefined location. Then it finds the best matching positive samples by the Hungarian algorithm. The traditional bipartite matching process is based on anchors \cite{DETR_introduced_bipartite_label_assignment}, while DETR \cite{DETR} leveraging object query changes this matching art to a total end-to-end style. Using the Hungarian algorithm, DETR computes loss directly between predefined object queries and ground-truth object boxes. Then after training, the converged queries can be directly decoded into the final predicted boxes without Non-Maximum Suppression (NMS). However, the size of the query map in our proposed method is set to be the same as the size of the radar and LiDAR feature map. Such a large number of queries leads to massive computational consumption, which is not feasible when applying the Hungarian algorithm. Thus, such bipartite matching label assignment is not considered for our proposed method.

\begin{figure*}[htbp]
    \centering
    \includegraphics[scale=0.67]{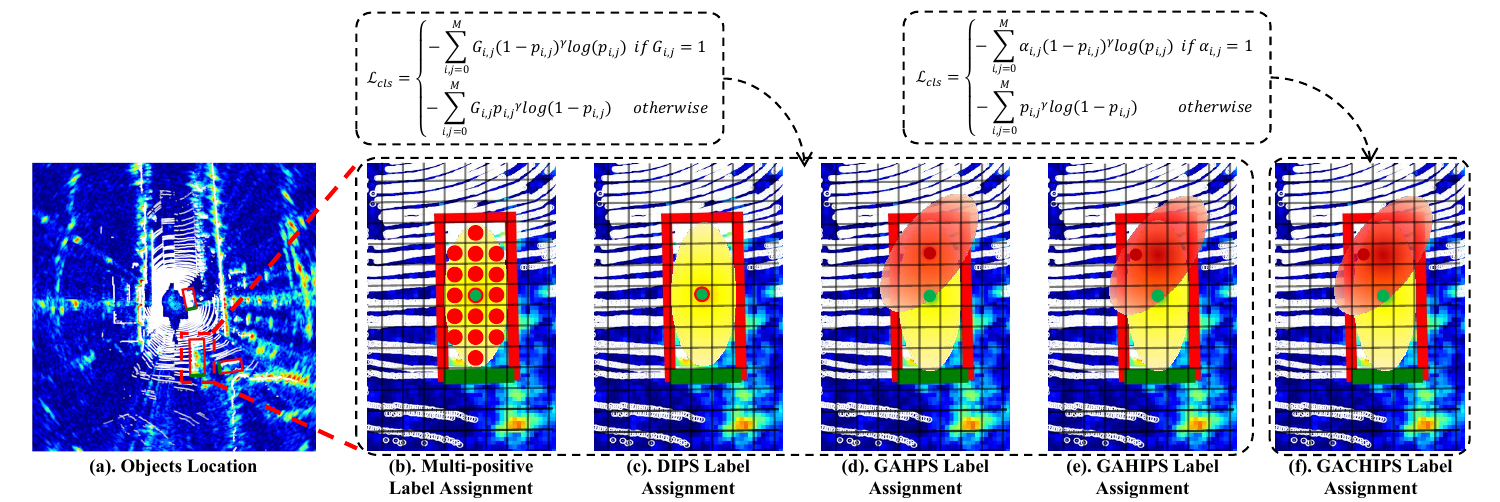}
    \caption{Overview of label assignment strategies for object detection. (a) Identifies the object with a red-bordered ground-truth bounding box and a green headline. Yellow and red ellipses represent ground-truth and predicted Gaussian distributions, respectively. Red dots are positive sample points, and green dots mark the ground-truth Gaussian centers. Strategies (b) to (e) show different methods for selecting positive samples for box loss calculation, ranging from (b) using all anchor points within the ground-truth area, to (c) selecting the center, (d) the point with the highest foreground-background classification score, and (e) the point with the highest "score plus IoU". Strategy (f) integrates the approach from (e) with an alternative loss function as Eq. (\ref{focal_loss_modifed_in_GACHIPS}).}
    \label{label assignment strategy figure}
\end{figure*}

\subsubsection{Multi-Positives Label Assignment}

Following the idea that multiple positive anchor points may alleviate positive and negative sample imbalance problems in anchor-free detection architecture, a Gaussian area-based multi-positive label assignment is designed. Based on the ground-truth Gaussian mixture distribution heatmap $\textbf{G}$ created in Section \ref{Description of the overall network}, a static threshold is used to extract the specific region as the positive anchor points. Foreground-background classification and box regression are then applied together on each positive anchors points. As Fig. \ref{label assignment strategy figure}(b) shows, all dots with red colors are the selected positive samples. They are all located in the ground-truth Gaussian distribution yellow-colored area, thus this approach is named as \emph{Gaussian Area-based Multi-Positives Sample assignment}.

Nonetheless, the introduction of multiple positive samples inevitably leads to competition in training procedure among them because they all devote to the gratitude, exacerbating the consistency gap during the interference phase. The experiment result is shown later in Section \ref{Ablation_Study_on_Label_Assignment_Strategies}.

\subsubsection{Single Positive Label Assignment}
% 为了避免多正样本带来的数值回归冲突，单正样本分配策略被设计。单正样本的基础特点是针对每一个真值目标，有且仅有一个正样本锚点与之对应，在每次训练中对loss贡献梯度。(正样本对梯度有贡献)
In order to prevent numerical regression conflicts caused by multiple positive samples, single positive sample allocation strategy has been designed. The fundamental characteristic of a single positive sample is that, for each ground-truth object, there is only one positive anchor point corresponding to it, which contributes to the gradient of the loss in each training iteration.
% 针对上述讨论中提到的前景背景分类与框回归存在的不一致性，四种方法被依次提出来缓解这个问题。
Four methods have been proposed successively to alleviate the inconsistency between foreground-background classification and box regression mentioned in the previous discussion. 
% 实验表明，抛弃中心先验，能够最有效地缓解目标的检测和回归的一致性问题。
The experiment shows that abandoning the center prior can most effectively alleviate the consistency issue between object detection and regression. The proposed four methods 
%are \emph{Direct Index based Positive Sample (DIPS) assignment}, \emph{Gaussian Area-based Heatmap cost Positive Sample (GAHPS) assignment}, \emph{Gaussian Area-based Heatmap and IoU cost Positive Sample (GAHIPS) assignment} and \emph{Gaussian Area-based Consistent Heatmap and IoU cost Positive Sample (GACHIPS) assignment}. They 
share the same training framework as below:

\textcircled{1}  Generate candidate anchor points: For each ground-truth object, create a set of candidate anchor points.

\textcircled{2} Design unique costs: Assign a unique cost to each candidate anchor point.

\textcircled{3} Select positive samples: Choose the most suitable candidate anchor points as positive samples.

\textcircled{4} Compute loss: Calculate the loss result based on the selected positive anchor points.

The four label assignment strategies contribute different innovations in the pipeline. The details are elaborated below:

The first step is generating candidate anchor points. This step is the common procedure for all four label assignment strategies. According to Eq. (\ref{gaussian_map_generation_equation}), a Gaussian distribution \{$\bm{G}_k, k = 1, ..., n$\} is generated for each ground-truth object, and the multiple objects form a Gaussian mixture distribution $\bm{G}$. By setting a fixed threshold, the anchor points inside each Gaussian distribution can be obtained, which are called candidate anchor points. This steps can be described as:

\begin{equation}
    \boldsymbol{A^G}_k = \boldsymbol{\mathcal{A}}\cap \boldsymbol{G}_k,
\end{equation}
where $\boldsymbol{A^G}_k$ represents the set of candidate anchor points in the $k$-th ground-truth Gaussian distribution, $\boldsymbol{\mathcal{A}}$ represents the set of all anchor points on the BEV grid, and $\cap$ represents the intersection operation.

The second and third steps are to design unique cost for these candidate anchor points and then select one as positive sample. In this two steps, the first three label assignment strategies make improvements rely on different design:

\begin{itemize}

\item The first design is \emph{Direct Index based Positive Sample (DIPS) assignment}. As Fig. \ref{label assignment strategy figure}(c) shows, DIPS treats the label center locations as the positive anchor points, then applies foreground-background classification and box regression on these anchor points. The DIPS sample assignment is a standard design of CenterPoint \cite{LiDAR_object_detection_CenterPoint} whose cost and selection are described as:

\begin{equation}
    \label{DIPS_compute_centers}
    a_k=computeCenter(\boldsymbol{A^G}_k).
\end{equation}

The process begins by determining if a candidate anchor point is positioned at the center of $\boldsymbol{A^G}_k$. This is calculated and represented as $a_k$ according to Eq. (\ref{DIPS_compute_centers}). Following this, a binary mask, symbolized as $\boldsymbol{\alpha}$, is used to indicate the selection of positive samples. This mask is assigned a value of 1 in accordance with: 
\begin{equation}
    \label{compute_alpha}
    \boldsymbol{\alpha}(a_k)=1.
\end{equation}
After the establishment of $\boldsymbol{\alpha}$, the location for the box regression loss is pinpointed. Subsequently, the final loss is then calculated based on Eq. (\ref{loss_formulation}), which is summation of two values obtained from Eq. (\ref{focal_loss}) and Eq. (\ref{box_loss}) for all the positive samples.

\item The second approach is termed as the \emph{Gaussian Area-based Heatmap cost Positive Sample (GAHPS) assignment}. In this method, we substitute the cost calculation from determining the centers of ground-truth boxes to assessing the maximum predicted heatmap value for each anchor point. The remaining steps, including the first and the fourth, are kept unchanged. Firstly, the model inferences predicted heatmap $\boldsymbol{P}^{cls}_k$ and boxes $\boldsymbol{P}^{box}_k$ follows Eq. (\ref{model_predicting}). As illustrated in Fig. \ref{label assignment strategy figure}(d), the red fading ellipse represents the predicted Gaussian distribution heatmap. This step is described as:

\begin{equation}
    \label{model_predicting}
    \boldsymbol{P}^{cls}_k, \boldsymbol{P}^{box}_k=Decode(\boldsymbol{F},\boldsymbol{A^G}_k).
\end{equation}

The decoding function, denoted as $Decode(\cdot)$, takes two inputs: the feature map $\boldsymbol{F}$ and the anchor points in $k$-th Gaussian distribution $\boldsymbol{A}^G_k$. Then, different from DIPS, the cost is designed as predicted heatmap as shown in: \begin{equation}
    \label{GAHPS_cost}
    \boldsymbol{C}_k^a=\boldsymbol{P}^{cls}_k.
\end{equation}
The anchor point with highest predicted heatmap value are chosen as the positive sample $a_k$ as shown in:
\begin{equation}
    \label{select_peak_anchor}
    a_k=arg max(\boldsymbol{C}_k^a).
\end{equation}

The Eqs. (\ref{model_predicting}), (\ref{GAHPS_cost}), and (\ref{select_peak_anchor}) are utilized to substitute Eq. (\ref{DIPS_compute_centers}). Following this, the binary mask $\boldsymbol{\alpha}$ is computed using Eq. (\ref{compute_alpha}). Similar to the first DIPS strategy, the final loss is also calculated using Eq. (\ref{loss_formulation}), (\ref{focal_loss}) and (\ref{box_loss}). However, the final loss is determined based on the established $\boldsymbol{\alpha}$ here in this design.

% 试图让回归参与进样本分配中，使得分类和回归同时作用于loss，来引导模型训练。
\item The third approach is termed as the \emph{Gaussian Area-based Heatmap and IoU cost Positive Sample (GAHIPS) assignment}, as shown in Fig. \ref{label assignment strategy figure}(e). In this method, we go beyond merely substituting the cost with the predicted heatmap value. Instead, we replace it with the sum of the predicted heatmap value and the IoU value between the predicted box and the ground-truth box. Therefore, the first step is to calculate the IoU score of predicted bounding box and ground-truth box. The decoded $\boldsymbol{P}^{box}_k$ derived from Eq. (\ref{model_predicting}) is utilized to compute the IoU scores $\boldsymbol{S}^{IoU}_k$ in relation to the ground-truth bounding boxes $\boldsymbol{\mathcal{G}}^{box}_k$: %Eq. (\ref{GAHIPS_computeIoU})

\begin{equation}
    \label{GAHIPS_computeIoU}
    \boldsymbol{S}^{IoU}_k = computeIoU(\boldsymbol{P}^{box}_k, \boldsymbol{\mathcal{G}}^{box}_{k}).
\end{equation}
Following this, the sum of the predicted heatmap values and IoU scores, represented as $\boldsymbol{C}^a_k$, is employed as the new cost, as indicated in: %Eq. (\ref{GAHIPS_cost}).

\begin{equation}
    \label{GAHIPS_cost}
    \boldsymbol{C}^{a}_k = \boldsymbol{P}^{cls}_k+\boldsymbol{S}^{IoU}_k.
\end{equation}

The positive anchor point $a_k$ is then selected using Eq. (\ref{select_peak_anchor}), and the binary mask $\boldsymbol{\alpha}$ is also determined by Eq. (\ref{compute_alpha}). Lastly, same as aforementioned strategies, the final loss is still calculated using Eq. (\ref{loss_formulation}), which is a combination of Eq. (\ref{box_loss}) and Eq. (\ref{focal_loss}).

To be noted that, although the processes of determining $\boldsymbol{\alpha}$ are different in DIPS, GAHPS and GAHIPS, the ways of computing final loss are same. However, because the foreground-background classification loss and box regression loss are independent from each other,
such definition of final loss leads to a consistency gap between the two loss above. To address this issue, the fourth design is proposed.

% 当分类和回归都采用同一个正样本对loss做贡献时，这个时候并不要求正样本点一定要处于目标的中心位置，而是随着训练自行寻找到最佳的位置

\item The fourth design is \emph{Gaussian Area-based Consistent Heatmap and IoU cost Positive Sample (GACHIPS) assignment}, which is a divergence of the GAHIPS. 
GACHIPS follows the previous three steps as the GAHIPS, but modifies the final step of loss computation as indicated in Fig. \ref{label assignment strategy figure}(f).
%
% All the previous designs conceal a fact that the positive sample position of foreground-background classification is designed to converge to the center of the ground-truth box. 
% %
% Meanwhile, the box regression positive sample position moves with the position of foreground-background classification when training step by step. Those strategies build a weak relation between the two losses. 
% %
% However, the position of positive samples is not necessarily required to be at the center of the ground-truth bounding box.
%
By doing it, the foreground-background classification loss and box regression loss are both controlled by $\bm{\alpha}$. Specifically, the former foreground-background classification loss function which is shown in Eq. (\ref{focal_loss}), should be replaced by: %Eq. (\ref{focal_loss_modifed_in_GACHIPS}):

\begin{equation}
    %\footnotesize
    \label{focal_loss_modifed_in_GACHIPS}
    \mathcal{L}_{cls} = \left\{
    \begin{aligned}
    &-\sum_{i,j=0}^M \textbf{$\alpha$}_{i,j}(1-p_{i,j})^\gamma log(p_{i,j})~{\rm if}~\textbf{$\alpha$}_{i,j}=1, \\
    &-\sum_{i,j=0}^M p_{i,j}^\gamma log(1-p_{i,j})~~~~~~~~~~\rm{otherwise}.
    \end{aligned}
    \right.
\end{equation}

\end{itemize}

%In the end, a toy visualization of aforementioned label assignment strategies is shown in Fig. \ref{label assignment strategy figure}.
%We also describe all the label assignment strategies and the corresponding training process in Algorithms \ref{algorithm_label_assign} and \ref{supervised_learning}.

During inference procedure, the key step is to extract peaks from the predicted heatmap. We set a static threshold to pick every anchor point to generate the corresponding box, then NMS is applied to discard other boxes that overlap with the higher confidence box by a certain size. Beware that NMS can still filter out a certain amount of overlapping bounding boxes estimated from three branches in multiple scales. However, the estimated bounding boxes rarely overlap in a single scale branch since the anchor box-free approach with a single positive label assignment strategy is employed.

% \iffalse
% \begin{figure*}
%     \centering
%     \includegraphics[scale=0.55]{Interactive Transformer Fusion.pdf}
%     \caption{Interactive Transformer BEV Fusion Approaches}
%     \label{Interactive Transformer Fusion Approaches}
% \end{figure*}
% \fi

% \iffalse
% \begin{figure*}
%     \centering
%     \includegraphics[scale=0.55]{Interactive Transformer Fusion_2.pdf}
%     \caption{Interactive Transformer BEV Fusion Approaches}
%     \label{Interactive Transformer Fusion Approaches}
% \end{figure*}
% \fi

\iffalse
\begin{figure*}[]
    \centering
    \includegraphics[width=0.9\textwidth]{Interactive Transformer Fusion TIV v8 wide version.pdf}
    \caption{Interactive transformer-based BEV fusion approaches.}
    \label{Interactive Transformer Fusion Approaches}
\end{figure*}
\fi

\begin{figure}[t]
    \centering
    \includegraphics[width=0.45\textwidth]{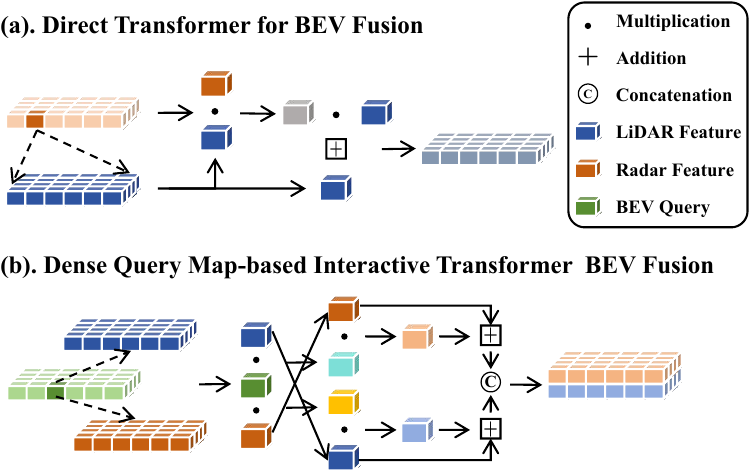}
    \caption{Comparison of direct transformer for BEV fusion with dense query map-based interactive transformer for BEV fusion.}
    \label{Interactive Transformer Fusion Approaches}
\end{figure}

% \iffalse
% \begin{figure}[ht]
%     \centering
%     \includegraphics[width=0.45\textwidth]{Interactive_Transformer_Fusion_3.pdf}
%     \caption{Interactive transformer BEV fusion approaches.}
%     \label{Interactive Transformer Fusion Approaches}
% \end{figure}
% \fi

\subsection{Interactive Transformer for Fusion of Radar and LiDAR in BEV}
In the pursuit of enhancing the performance of sensor fusion in autonomous driving systems, we propose a novel approach that transcends the conventional method of simply concatenating features from each sensor. The standard transformer \cite{self_attention} based traditional approach, while straightforward, is inherently limited in its ability to effectively weigh the features from each sensor. To this end, we have designed an innovative interactive transformer-based BEV Fusion block, to fuse data from the two sensor branches more effectively.

The first approach \emph{Direct Transformer for BEV Fusion}, depicted in Fig. \ref{Interactive Transformer Fusion Approaches}(a), employs a plain transformer BEV fusion as a basic method. This method utilizes radar branch feature map as the query and LiDAR feature map as the key and value. The weight matrix is computed through matrix multiplication of the query and key, which is subsequently normalized using a softmax function. Then the dot product of the weight and value are added with value itself acting as a residual process. This process can be mathematically represented as follows:
\begin{equation}
    \label{direct_interactive_transformation}
    \bm{X}_{out}=\sigma(\bm{X}_{R}^{}\bm{X}_{L}^T)\odot\bm{X}_{L}+\bm{X}_{L}.
\end{equation}
In this equation, $\bm{X}$ represents the feature map, $\sigma(\cdot)$ denotes the softmax function, $\odot$ denotes element-wise multiplication, and {$\bm{X}_R$, $\bm{X}_L$} are radar and LiDAR feature maps. This method serves as a baseline to demonstrate the effectiveness of our second approach.

The second approach, shown in Fig. \ref{Interactive Transformer Fusion Approaches}(b), named \emph{Dense Query Map-based Interactive Transformer for BEV Fusion (DQMITBF)}, shares the core concept with the first approach but introduces a variation in the query definition. In this method, we define a randomly initialized learnable BEV query that shares the same size as the input key feature map. Interactive attention is computed between this learnable BEV query and both the radar and LiDAR feature maps, which are treated as the key and value, respectively. The interactive fusion of LiDAR and radar features through a BEV query involves a three-step process. Firstly, to facilitate radar-to-LiDAR interaction fusion, the BEV query is used to calculate the correlation between itself and the radar feature map. Secondly, the resulting softmax output is then utilized as a weighting factor for the LiDAR feature map. Finally, the weighted LiDAR feature map is added to the LiDAR feature itself. Similarly, the same fusion procedure is repeated for LiDAR-to-radar features. In the end, the fused radar and LiDAR feature maps are concatenated for further processing. The fusion process can be succinctly represented as Eq. (\ref{query_based_transformer_equation}). In this equation, $\bm{X}_{Q}$ is the randomly initialized learnable query, $\bm{X}_R$ and $\bm{X}_L$ are the feature maps of radar and LiDAR, respectively. $\textcircled{c}$ is concatenation process.

\begin{figure*}
\begin{equation}
\begin{aligned}
\label{query_based_transformer_equation}    
\bm{X}_{out} = (\sigma(\bm{X}_{Q}^{}\bm{X}_{R}^T)\odot{\bm{X}_{L}}+\bm{X}_{L}) \textcircled{c} (\sigma(\bm{X}_{Q}^{}\bm{X}_{L}^T)\odot{\bm{X}_{R}}+\bm{X}_{R}).
\end{aligned}
\end{equation}
\end{figure*}

The DQMITBF architecture interacts symmetrically with radar and LiDAR features through a BEV query and integrates them in a unified phase. This interactive query allows the model to prioritize feature-rich branches, especially in varying weather conditions. Thus, this novel approach optimizes the weighting and combination of heterogeneous sensor features in an adaptive manner, enhancing the effectiveness of autonomous driving systems in different meteorological scenarios.

%The third fusion fashion in Fig. \ref{Interactive Transformer Fusion Approaches}(c) is query-based dual interactive BEV fusion. The pre-defined query is also required to communicate with both the LiDAR and radar feature map. So here we use concatenated LiDAR and radar feature map as key. It's worthy to mention that because the query can not match the same C dimension, the actual attention method is to do the attention with each feature map respectively. The equation 7 can describe it.
%\begin{equation}
%    \bm{X}_{out}=\sigma(\bm{X}_{Q}^{}\bm{X}_{C}^T)%\odot{\bm{X}_C}+\bm{X}_{C} 
%\end{equation}
%and
%\begin{equation}
%    \bm{X}_{C}=\bm{X}_{s_0}\oplus\bm{X}_{s_1}
%\end{equation}
%where $\odot$ and $\oplus$ denote the element-wise multiplication and concatenation operation, respectively.

%The last fusion logic applies deformable attention. to be continued ...

\section{Implementation and Results}\label{sec results}
\subsection{Dataset and Evaluation Metrics}

The ORR dataset \cite{Oxford_Robotcar_Dataset} is selected because it provides a synchronized LiDAR point cloud, radar range azimuth heatmap, car motion pose parameters, etc. The LiDAR point cloud was captured by two Velodyne HDL-32E LiDARs, each designed to cover one side of the car. On the top of the car locates a NavTech CTS350-X radar, which has a $360^{\circ}$ field of view in the fashion of a mechanical scan. The angle resolution of radar data is $0.9^{\circ}$ every 400ms, while LiDAR data angle resolution is $0.33^{\circ}$ every 20ms.

The synchronization issue between LiDAR data and radar data and the annotation generation was solved by interpolation with the help of ego vehicle pose \cite{radar_LiDAR_fusion_object_detection_MVDNet}. Specifically, the pose parameters compensated for the misalignment between the LiDAR point cloud and radar range-azimuth heatmap. Utilizing the timestamps as a reference, \cite{radar_LiDAR_fusion_object_detection_MVDNet} computes the motion between LiDAR point cloud and radar heatmap, then transfer every point to radar coordinate. The annotation bounding boxes are also created by \cite{radar_LiDAR_fusion_object_detection_MVDNet} every 20 frames. The remaining frames' annotations were also generated by interpolation. For testing in a foggy scenario, MVDNet relocates the LiDAR point cloud with the method of DEF \cite{DEF}. According to the fog model in DEF, MVDNet sets the fog probability to 0.05 to recompute a maximum visible range for every point. If the current point range is greater than the maximum foggy range, this point would be set rather lost or relocated as a scatter point. Following the same operations on data in MVDNet, the 2D label bounding boxes are used as supervision. Corresponding radar and LiDAR data in 8862 number frames are split into 7071 for training and 1791 for testing. 

To evaluate the prediction result of the model, the 2D rotated bounding box average precision (AP) under different IoU thresholds is employed as a final evaluation metric. This metric simultaneously evaluates the precision and recall performance of the model under the test dataset. A higher AP value represents better prediction precision of the bounding box and a smaller missing detection rate of the objects. Prediction boxes can be defined as True Positive (TP) objects, which overlap with the ground-truth objects over the IoU threshold, or False Positive (FP) objects which overlap with ground-truth objects lesser than the IoU threshold. By setting different IoU thresholds, the number of TP and FP objects varies. Under a certain IoU threshold value, all the prediction boxes can be arranged as a TP/FP list sorted by their class scores. The precision and recall performance is defined as:
\begin{equation}
    \label{PR_definition}
    \begin{aligned}
    &Precision~=~\frac{TP}{TP+FP}, \\
    &Recall~~~~~=~\frac{TP}{TP+FN}.
    \end{aligned}
\end{equation}
The $TP+FP$ equals the total number of prediction objects, and the $FP$ is the number of false prediction objects. While the $FN$ is the number of miss-detection objects, $TP+FN$ equals the number of ground-truth objects. 

The Table \ref{pr_curve_list} shows an example of how the AP is calculated by setting the IoU threshold as 0.5. Assuming no miss-detected objects exist, all these predicted objects represented in the column are sorted by their scores. Precision and recall are computed according to Eq. (\ref{PR_definition}) in each column. It is worth mentioning that the precision result drops and the recall increase, along with the decrease of the object's scores. The AP@IoU=0.5 is then defined as the area value enclosed under the precision-recall curve accordingly to reflect the trend of the precision-recall curve as a performance metric. The greater the AP value is, the better model performs.
%Tao:下面一句话是想说什么？
% 杨炎龙：这句话主要是想说，像前面描述一样，得到了PR曲线之后，AP就是PR曲线的面积，而AP是我们最终的metric。
%Therefore, the Precision and Recall build a curve, and the AP is defined as the area enclosed in the pr curve.

\begin{table}[]
\caption{Example list of PR curve at IoU=0.5}
\label{pr_curve_list}
\begin{center}
\scalebox{0.95}{
\begin{tabular}{|c|c|c|c|c|c|c|c|}
\hline
Scores            & 0.98 & 0.97 & 0.96 & 0.95 & 0.94 & 0.92 & 0.91 \\ \hline
Prediction Objects & TP   & TP   & FP   & TP   & TP   & TP   & FP   \\ \hline
Precision          & 1    & 1    & 2/3  & 3/4  & 4/5  & 5/6  & 5/7  \\ \hline
Recall             & 1/5  & 2/5  & 2/5  & 3/5  & 4/5  & 1    & 1    \\ \hline
\end{tabular} }
\end{center}
\end{table}
%To evprediction aluate the boxes aprall the ediction result of model, the 2D rotated bounding box average precision(AP) is employed as final evaluation metrics. This metric requires a IoU threshold to define the prediction 2d box a truth positive target identity or a false alarm target comparing to the ground-truth bounding box at the same frame. Then count the proportion of all positive targets in the prediction result as the precision, the proportion of all positive targets in the ground-truth result as the recall. With the recall value increase, the precision value will drop, these two indicator build a curve. AP is the area under this curve. Since different IoU threshold creates different AP curve, the metric is usually described with a threshold icon such as AP@IoU=0.5, which means the AP under IoU threshold 50\%. The AP@IoU=0.5, AP@IoU=0.65 and AP@IoU=0.8 are chosen to evaluate the complete performance of each experiment. Besides, since this dataset only provides one object class, the AP is merely car class AP.

%\textcolor{blue}{(Same as MVDNet, COCO evaluator is used as the metrics method \textcolor{green}{This sentence needs to be re-written. We should not mention anything in the code, e.g. COCO evaluator, but need to use the formal name of evaluation metrics, e.g. AP@IoU = xxx, and introduce what is AP, how it has been calculated, etc.}. It evaluates the IoU between predicted bounding boxes and ground-truth under 0.5, 0.65, 0.8.\textcolor{(, as well as 0.5:0.05:0.95)} Since this dataset only provides one object class, the AP is merely car class AP.)}

\subsection{Implementing Details}

The preprocess of LiDAR data is based on PointPillars \cite{LiDAR_object_detection_PointPillar}, where we set $0.2m\times 0.2m$ as our BEV spatial resolution. It transforms the disordered LiDAR point cloud to a $320\times320\times9$ structured matrix. Meanwhile, the input radar data is in the format of $320\times320\times1$ under BEV, naturally. As Fig. \ref{RaLiBEV Framework} shows, the subsequent convolution block consists of a $3\times3$ convolution layer, a batch norm layer, and a leaky ReLu layer. The replication N of it is set as 3. The inference procedure decodes the predicted boxes from three scale output anchor-free heads. The static threshold is set to 0.1 to extract peaks (as key points among all anchor points) from the predicted heatmap. 
%Peak prune method is utilized for finding a more specific peak index, whose sliding window size is $3\times3$, the maximum pixel in this window will be chose.
After generating boxes from three scale output feature maps, the NMS method is applied for deleting extra boxes, whose threshold is set as 0.2.

When training, the batch size is set to 32, and 4 Nvidia Tesla V100S GPUs are used. The first two epochs are used as a warming up. The learning rate linearly increases from 0 to 0.01 after 2 epochs' iterations and then decreases by a ratio of 0.1 every 10 epochs. ADAM optimizer is used in the training process.

\subsection{Experimental Results and Analysis}

\begin{table*}[ht]
\caption{Ablation study of RaLiBEV anchor box free object detector.} % \vspace{-15pt}}
\label{ablation_study_ralibev}
\begin{center}
\scalebox{1.0}{
\begin{tabular}{|ccccc|}
\hline
%\multirow{2}\multicolumn{2}{|c|}{Method} 
\multicolumn{2}{|c|}{Methods}  & \multicolumn{1}{c|}{AP@IoU=0.5}  & \multicolumn{1}{c|}{AP@IoU=0.65} & \multicolumn{1}{c|}{AP@IoU=0.8} %&\multicolumn{1}{|c|}{inference time(s)} 
\\ \hline
%\multicolumn{5}{|c|}{Classical LiDAR only Method as \textbf{Baselines}}
%\\ \hline
%\multicolumn{2}{|c|}{PointPillar (CVPR 2019)}                               & \multicolumn{1}{c|}{85.8}     & \multicolumn{1}{c|}{82.9}     & \multicolumn{1}{c|}{60.6}     %&\multicolumn{1}{c|}{} 
%\\ \hline
%\multicolumn{5}{|c|}{State-of-the-art Radar and LiDAR Fusion Methods as \textbf{Baselines}}
%\\ \hline
%\multicolumn{2}{|c|}{DEF (CVPR 2020)}                            & \multicolumn{1}{c|}{85.9}     & \multicolumn{1}{c|}{78.1}     & \multicolumn{1}{c|}{44.2}      %&\multicolumn{1}{c|}{}      
%\\ \hline
%\multicolumn{2}{|c|}{MVDNet (CVPR 2021)}                               & \multicolumn{1}{c|}{87.2}     & \multicolumn{1}{c|}{86.1}     & \multicolumn{1}{c|}{72.6}     %&\multicolumn{1}{c|}{} 
%\\ \hline
%\multicolumn{2}{|c|}{ST-MVDNet (CVPR 2022)}                            & \multicolumn{1}{c|}{91.4}     & \multicolumn{1}{c|}{89.9}     & \multicolumn{1}{c|}{78.4}     %&\multicolumn{1}{c|}{}       
%\\ \hline
\multicolumn{5}{|c|}{Proposed RaLiBEV with Different Label Assignments and Plain BEV Fusion (\textbf{Ours})}
\\ \hline
%\multicolumn{2}{|c|}{(1)Hungarian Algorithm-based centers distance cost Positive Sample Assignment(x)}            & \multicolumn{1}{c|}{93.0} & \multicolumn{1}{c|}{89.2} & \multicolumn{1}{c|}{74.9}  &\multicolumn{1}{c|}{0.674}         \\ \hline
\multicolumn{2}{|c|}{%(3)
Gaussian Area-based Multi-Positives Sample Assignment}            & \multicolumn{1}{c|}{95.4} & \multicolumn{1}{c|}{94.2} & \multicolumn{1}{c|}{75.7}  %&\multicolumn{1}{c|}{0.753}     
\\ \hline
\multicolumn{2}{|c|}{%(2)
Direct Index-based Positive Sample assignment (DIPS)}            & \multicolumn{1}{c|}{96.3} & \multicolumn{1}{c|}{94.9} & \multicolumn{1}{c|}{84.5}  %&\multicolumn{1}{c|}{0.674}       
\\ \hline
\multicolumn{2}{|c|}{%(5)
Gaussian Area-based Heatmap cost Positive Sample assignment (GAHPS)}            & \multicolumn{1}{c|}{96.7} & \multicolumn{1}{c|}{95.4} & \multicolumn{1}{c|}{89.5}  %&\multicolumn{1}{c|}{0.669}        
\\ \hline
\multicolumn{2}{|c|}{%(6)
Gaussian Area-based Heatmap and IoU cost Positive Sample assignment (GAHIPS)}            & \multicolumn{1}{c|}{96.8} & \multicolumn{1}{c|}{96.7} & \multicolumn{1}{c|}{92.0}  %&\multicolumn{1}{c|}{0.662}         
\\ \hline
\multicolumn{2}{|c|}{%(7)
Gaussian Area-based Consistent Heatmap and IoU cost Positive Sample assignment (GACHIPS)}            & \multicolumn{1}{c|}{\textbf{97.4}} & \multicolumn{1}{c|}{\textbf{96.7}} & \multicolumn{1}{c|}{\textbf{93.9}}  %&\multicolumn{1}{c|}{0.719}         
\\ \hline
\multicolumn{5}{|c|}{Proposed RaLiBEV with Interactive Transformer-based BEV Fusion (\textbf{Ours})}           
\\ \hline
\multicolumn{2}{|c|}{Direct Transformer for BEV Fusion %(with DIPSA)
}                               & \multicolumn{1}{c|}{97.8} & \multicolumn{1}{c|}{97.5} & \multicolumn{1}{c|}{92.3}  %&\multicolumn{1}{c|}{0.708}         
\\ \hline
%\multicolumn{2}{|c|}{Direct Interactive BEV Fusion, Advanced(1.2)}                               & \multicolumn{1}{c|}{95.8 / 97.9}     & \multicolumn{1}{c|}{94.7 / 97.7}     & \multicolumn{1}{c|}{89.5 / 93.5} &\multicolumn{1}{c|}{0.763}        \\ \hline
%\multicolumn{2}{|c|}{Query-based Interactive BEV Fusion, Basic(2.1)}                               & \multicolumn{1}{c|}{/ 96.2}     & \multicolumn{1}{c|}{/ 94.7}     & \multicolumn{1}{c|}{/ 82.0}     &\multicolumn{1}{c|}{0.703} \\ \hline
%\multicolumn{2}{|c|}{Query-based Interactive BEV Fusion, Advanced((2.2))}                               & \multicolumn{1}{c|}{97.9}     & \multicolumn{1}{c|}{97.8}     & \multicolumn{1}{c|}{93.9}  &\multicolumn{1}{c|}{0.773} \\ \hline
\multicolumn{2}{|c|}{Dense Query Map-based Interactive Transformer for BEV Fusion (DQMITBF) %(with DIPS)
}                               & \multicolumn{1}{c|}{\textbf{98.8}}     & \multicolumn{1}{c|}{\textbf{97.7}}     & \multicolumn{1}{c|}{\textbf{94.8}}   %&\multicolumn{1}{c|}{0.678} 
\\ \hline
%\multicolumn{2}{|c|}{Direct Interactive BEV Fusion, with GACHIPS}                               & \multicolumn{1}{c|}{97.7} & \multicolumn{1}{c|}{96.7} & \multicolumn{1}{c|}{93.9}  %&\multicolumn{1}{c|}{}        
%\\ \hline
%\multicolumn{2}{|c|}{Dense Query Map-based Interactive BEV Fusion, with GACHIPS}                               & \multicolumn{1}{c|}{97.8}     & \multicolumn{1}{c|}{96.8}     & \multicolumn{1}{c|}{93.9}   %&\multicolumn{1}{c|}{} 
%\\ \hline
%\multicolumn{2}{|c|}{Direct Interactive BEV Fusion with Deformable Attention}           & \multicolumn{1}{c|}{}     & \multicolumn{1}{c|}{}     & \multicolumn{1}{c|}{}      %&\multicolumn{1}{c|}{} 
%\\ \hline
%\multicolumn{2}{|c|}{Dense Query Map-based Interactive BEV Fusion with Deformable Attention}           & \multicolumn{1}{c|}{}     & \multicolumn{1}{c|}{}     & \multicolumn{1}{c|}{}      %&\multicolumn{1}{c|}{} 
%\\ \hline
\end{tabular}
}
\end{center}
\end{table*}

\begin{table*}[ht]
\caption{Comparison of LiDAR only approaches and RaLiBEV which employs fusion of radar and LiDAR. The numbers in bold are the best of all methods.}
\label{radar_lidar_vs_lidar_only}
\begin{center}
\scalebox{1.05}{
\begin{tabular}{|cc|c|cccccc|cccccc|}
\hline
\multicolumn{1}{|c|}{\multirow{3}{*}{Method}} & Train & \multirow{3}{*}{Modality} & \multicolumn{6}{c|}{Clear+Foggy}                                                                                                                 & \multicolumn{6}{c|}{Clear-only}                                                                                                                  \\ \cline{2-2} \cline{4-15} 
\multicolumn{1}{|c|}{}                        & Test  &                           & \multicolumn{3}{c|}{Clear}                                                        & \multicolumn{3}{c|}{Foggy}                                   & \multicolumn{3}{c|}{Clear}                                                        & \multicolumn{3}{c|}{Foggy}                                   \\ \cline{2-2} \cline{4-15} 
\multicolumn{1}{|c|}{}                        & IoU   &                           & \multicolumn{1}{c|}{0.5}  & \multicolumn{1}{c|}{0.65} & \multicolumn{1}{c|}{0.8}  & \multicolumn{1}{c|}{0.5}  & \multicolumn{1}{c|}{0.65} & 0.8  & \multicolumn{1}{c|}{0.5}  & \multicolumn{1}{c|}{0.65} & \multicolumn{1}{c|}{0.8}  & \multicolumn{1}{c|}{0.5}  & \multicolumn{1}{c|}{0.65} & 0.8  \\ \hline
\multicolumn{2}{|c|}{PointPillars}        & L                         & \multicolumn{1}{c|}{85.8} & \multicolumn{1}{c|}{83.0} & \multicolumn{1}{c|}{58.3} & \multicolumn{1}{c|}{72.8} & \multicolumn{1}{c|}{70.3} & 48.6 & \multicolumn{1}{c|}{85.8} & \multicolumn{1}{c|}{82.9} & \multicolumn{1}{c|}{60.6} & \multicolumn{1}{c|}{71.3} & \multicolumn{1}{c|}{68.3} & 47.8 \\ \hline
\multicolumn{2}{|c|}{RaLiBEV LiDAR-Only}                         & L                         & \multicolumn{1}{c|}{96.3}     & \multicolumn{1}{c|}{93.8}     & \multicolumn{1}{c|}{78.3}     & \multicolumn{1}{c|}{92.2}     & \multicolumn{1}{c|}{88.3}     &   74.4   & \multicolumn{1}{c|}{97.7}     & \multicolumn{1}{c|}{96.4}     & \multicolumn{1}{c|}{87.5}     & \multicolumn{1}{c|}{68.2}     & \multicolumn{1}{c|}{62.7}     &   40.2   \\ \hline
% \multicolumn{2}{|c|}{RaLiBEV Radar-Only}                         & R                         & \multicolumn{1}{c|}{97.7}     & \multicolumn{1}{c|}{96.5}     & \multicolumn{1}{c|}{89.6}     & \multicolumn{1}{c|}{97.7}     & \multicolumn{1}{c|}{96.5}     &   89.6   & \multicolumn{1}{c|}{97.7}     & \multicolumn{1}{c|}{96.5}     & \multicolumn{1}{c|}{89.6}     & \multicolumn{1}{c|}{\textbf{97.7*}}     & \multicolumn{1}{c|}{\textbf{96.5*}}     &   \textbf{89.6*}   \\ \hline
\multicolumn{2}{|c|}{RaLiBEV with GACHIPS (Ours)}     & L+R                       & \multicolumn{1}{c|}{97.8} & \multicolumn{1}{c|}{97.7} & \multicolumn{1}{c|}{93.9} & \multicolumn{1}{c|}{97.8} & \multicolumn{1}{c|}{96.7} & 93.7 & \multicolumn{1}{c|}{97.4} & \multicolumn{1}{c|}{96.7} & \multicolumn{1}{c|}{93.9} & \multicolumn{1}{c|}{\textbf{84.6}} & \multicolumn{1}{c|}{\textbf{82.4}} & \textbf{73.8} \\ \hline
\multicolumn{2}{|c|}{RaLiBEV with DQMITBF (Ours)}     & L+R                       & \multicolumn{1}{c|}{\textbf{98.8}} & \multicolumn{1}{c|}{\textbf{97.7}} & \multicolumn{1}{c|}{\textbf{95.1}} & \multicolumn{1}{c|}{\textbf{97.9}} & \multicolumn{1}{c|}{\textbf{97.8}} & \textbf{94.1} & \multicolumn{1}{c|}{\textbf{98.8}} & \multicolumn{1}{c|}{\textbf{97.7}} & \multicolumn{1}{c|}{\textbf{94.8}} & \multicolumn{1}{c|}{83.3} & \multicolumn{1}{c|}{80.9} & 70.6 \\ \hline
\end{tabular}
}
\end{center}
\end{table*}

%For foggy LiDAR
\begin{table*}[ht]
\caption{Comparison of AP of oriented bounding boxes in bird's eye view between our proposed RaLiBEV and other state-of-the-art radar and LiDAR fusion-based methods. The numbers in bold are the best of all methods.}
\label{sota_comparison}
\begin{center}
\scalebox{1.05}{
\begin{tabular}{|cc|c|cccccc|cccccc|}
\hline
\multicolumn{1}{|c|}{\multirow{3}{*}{Method}} & Train & \multirow{3}{*}{Modality} & \multicolumn{6}{c|}{Clear+Foggy}                                                                                                                 & \multicolumn{6}{c|}{Clear-only}                                                                                                                  \\ \cline{2-2} \cline{4-15} 
\multicolumn{1}{|c|}{}                        & Test  &                           & \multicolumn{3}{c|}{Clear}                                                        & \multicolumn{3}{c|}{Foggy}                                   & \multicolumn{3}{c|}{Clear}                                                        & \multicolumn{3}{c|}{Foggy}                                   \\ \cline{2-2} \cline{4-15} 
\multicolumn{1}{|c|}{}                        & IoU   &                           & \multicolumn{1}{c|}{0.5}  & \multicolumn{1}{c|}{0.65} & \multicolumn{1}{c|}{0.8}  & \multicolumn{1}{c|}{0.5}  & \multicolumn{1}{c|}{0.65} & 0.8  & \multicolumn{1}{c|}{0.5}  & \multicolumn{1}{c|}{0.65} & \multicolumn{1}{c|}{0.8}  & \multicolumn{1}{c|}{0.5}  & \multicolumn{1}{c|}{0.65} & 0.8  \\ \hline
\multicolumn{2}{|c|}{DEF (CVPR2020) \cite{DEF}}                  & L+R                       & \multicolumn{1}{c|}{86.6} & \multicolumn{1}{c|}{78.2} & \multicolumn{1}{c|}{46.2} & \multicolumn{1}{c|}{81.4} & \multicolumn{1}{c|}{72.5} & 41.1 & \multicolumn{1}{c|}{85.9} & \multicolumn{1}{c|}{78.1} & \multicolumn{1}{c|}{44.2} & \multicolumn{1}{c|}{71.8} & \multicolumn{1}{c|}{63.7} & 32.4 \\ \hline
\multicolumn{2}{|c|}{MVDNet (CVPR2021) \cite{radar_LiDAR_fusion_object_detection_MVDNet}}               & L+R                       & \multicolumn{1}{c|}{90.9} & \multicolumn{1}{c|}{88.8} & \multicolumn{1}{c|}{74.6} & \multicolumn{1}{c|}{87.4} & \multicolumn{1}{c|}{84.6} & 68.9 & \multicolumn{1}{c|}{87.2} & \multicolumn{1}{c|}{86.1} & \multicolumn{1}{c|}{72.6} & \multicolumn{1}{c|}{78.0} & \multicolumn{1}{c|}{75.9} & 61.6 \\ \hline
\multicolumn{2}{|c|}{Bi-LRFusion (CVPR2023) \cite{radar_LiDAR_fusion_object_detection_Bi-LRFusion}}          & L+R                       & \multicolumn{1}{c|}{92.2} & \multicolumn{1}{c|}{N/A}  & \multicolumn{1}{c|}{74.4} & \multicolumn{1}{c|}{N/A}  & \multicolumn{1}{c|}{N/A}  & N/A  & \multicolumn{1}{c|}{N/A}  & \multicolumn{1}{c|}{N/A}  & \multicolumn{1}{c|}{N/A}  & \multicolumn{1}{c|}{N/A}  & \multicolumn{1}{c|}{N/A}  & N/A  \\ \hline
\multicolumn{2}{|c|}{ST-MVDNet (CVPR2022) \cite{radar_LiDAR_fusion_object_detection_ST-MVDNet}}            & L+R                       & \multicolumn{1}{c|}{94.7} & \multicolumn{1}{c|}{93.5} & \multicolumn{1}{c|}{80.7} & \multicolumn{1}{c|}{91.8} & \multicolumn{1}{c|}{88.3} & 73.6 & \multicolumn{1}{c|}{91.4} & \multicolumn{1}{c|}{89.9} & \multicolumn{1}{c|}{78.4} & \multicolumn{1}{c|}{81.2} & \multicolumn{1}{c|}{80.8} & 64.9 \\ \hline
\multicolumn{2}{|c|}{ST-MVDNet++ (ICASSP 2023) \cite{radar_LiDAR_fusion_object_detection_ST-MVDNet++}}       & L+R                       & \multicolumn{1}{c|}{96.0} & \multicolumn{1}{c|}{94.7} & \multicolumn{1}{c|}{82.0} & \multicolumn{1}{c|}{93.4} & \multicolumn{1}{c|}{90.0} & 75.1 & \multicolumn{1}{c|}{93.2} & \multicolumn{1}{c|}{92.0} & \multicolumn{1}{c|}{81.5} & \multicolumn{1}{c|}{83.7} & \multicolumn{1}{c|}{\textbf{83.2}} & 67.5 \\ \hline
\multicolumn{2}{|c|}{RaLiBEV with GACHIPS (Ours)}     & L+R                       & \multicolumn{1}{c|}{97.8} & \multicolumn{1}{c|}{97.7} & \multicolumn{1}{c|}{93.9} & \multicolumn{1}{c|}{97.8} & \multicolumn{1}{c|}{96.7} & 93.7 & \multicolumn{1}{c|}{97.4} & \multicolumn{1}{c|}{96.7} & \multicolumn{1}{c|}{93.9} & \multicolumn{1}{c|}{\textbf{84.6}} & \multicolumn{1}{c|}{82.4} & \textbf{73.8} \\ \hline
\multicolumn{2}{|c|}{RaLiBEV with DQMITBF (Ours)}     & L+R                       & \multicolumn{1}{c|}{\textbf{98.8}} & \multicolumn{1}{c|}{\textbf{97.7}} & \multicolumn{1}{c|}{\textbf{95.1}} & \multicolumn{1}{c|}{\textbf{97.9}} & \multicolumn{1}{c|}{\textbf{97.8}} & \textbf{94.1} & \multicolumn{1}{c|}{\textbf{98.8}} & \multicolumn{1}{c|}{\textbf{97.7}} & \multicolumn{1}{c|}{\textbf{94.8}} & \multicolumn{1}{c|}{83.3} & \multicolumn{1}{c|}{80.9} & 70.6 \\ \hline
\end{tabular}
}
\end{center}
\end{table*}

\begin{figure*}[htb]
    \centering
    \includegraphics[width=0.70\textwidth]{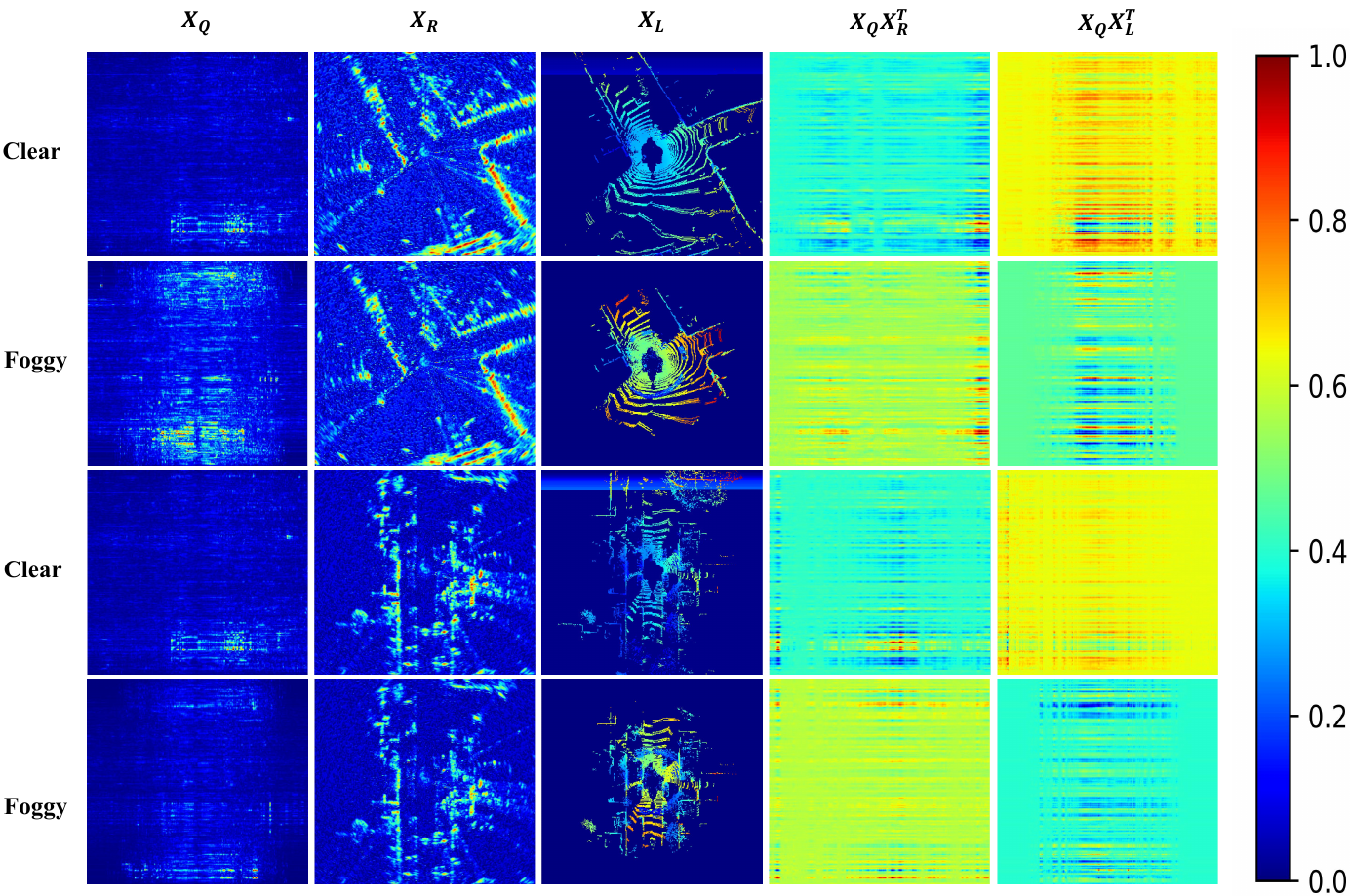}
    \caption{Interactive transformer feature map visualization results.}
    \label{Interactive transformer feature map visualization}
\end{figure*}

\subsubsection{Ablation Study on Label Assignment Strategies}
\label{Ablation_Study_on_Label_Assignment_Strategies}
Experiments on Oxford Radar RobotCar dataset are 
performed to evaluate the effectiveness of the proposed strategies. Since the annotation has only one type of labeled car class, the AP is used only for cars. The evaluation metrics of AP at IoU 0.5, 0.65, and 0.8 are chosen to demonstrate changes in performance.

Table \ref{ablation_study_ralibev} presents the experiment results of different label assignment strategies and interactive transformer fusion architectures of our proposed RaLiBEV. It is noted that the performance of the \textbf{Gaussian Area-based Multi-Positive Sample Assignment} strategy serves as a solid baseline thanks to the use of YOLOv4 backbone and neck in our RaLiBEV framework, which is similar to the core design of PillarNeXt \cite{LiDAR_object_detection_PillarNeXt}. The AP of IoU of 0.5 and 0.65 is 95.4\% and 94.2\%, both improve 2.2\%, but show a 5.8\% drop of AP at IoU of 0.8, falling down to 75.7\% compared to the ST-MVDNet++ \cite{radar_LiDAR_fusion_object_detection_ST-MVDNet++}. The AP performance of the state-of-the-art ST-MVDNet++ \cite{radar_LiDAR_fusion_object_detection_ST-MVDNet++} at IoU 0.5, 0.65, and 0.8 are 93.2\%, 92.0\%, and 81.5\%, respectively.
% 由于表1去掉了前面几个state-of-the-art的性能，故这里只能从表中有的信息来描述，可能稍微带一点与state-of-the-art相比的话。

The second type of label assignment strategy is \textbf{DIPS Assignment}. The improvement presented by this strategy is not significant at the low precision AP result at IoU of 0.5 and 0.65, approximately around 0.7\% - 0.9\%. Still, better performance at high precision AP at IoU of 0.8 results about 6\% compares to the first strategy above. The result of the third strategy illustrates the effectiveness of choosing a more appropriate positive-sample strategy. Compared to the previous strategy, the performance of \textbf{GAHPS} shows another performance improvement of about 5\% in high precision, which reaches 89.5\% AP at IoU of 0.8. Following this approach, in \textbf{GAHIPS}, the performance at an IoU of 0.5 remains almost the same as the previous strategy, while the most notable change is in AP at an IoU of 0.8. The performance breaks through to 92.0\%, which is a significant improvement of 10.5\% compared to the 81.5\% of the state-of-the-art ST-MVNDet++.

The above single positive label assignment strategies are based on encouraging the best box regression predicted key point to approach the classification predicted key point. These types of ideas have shown great validity. However, in the following strategy, a strong binding strategy has been experimented with. \textbf{GACHIPS} inherits the previous strategy, using heatmap and IoU as cost, while calculating foreground-background classification loss and box loss after finding the best-matched anchor point. The final result shows further improvement at AP of IoU 0.5 and 0.8, which increased to 97.4\% and 93.9\%, respectively.

% 实验分析：1. 在图像目标检测中设计的多正样本策略实际上是为了解决漏检测等问题，引入更多的正样本点使得模型更好地区分了目标和背景，而BEV视角下的检测问题中属于目标的bounding boxes并不会重叠，不存在由于目标重叠导致部分目标的pixels较少而难以识别为前景的问题，（因此引入多正样本的设计实际上对模型的预测精度带来了更大的问题）。而简单引入了多正样本未考虑其对回归带来的问题，那就是降低了模型的回归精度。实验结果表明，许多被认定为正样本的关键点的回归结果与真值框的重叠较小，它们处于远离真值框的中心点的位置，具有与真值框大小一致的预测框，但存在较大的偏移。这一点在统计指标上的表现则是，低精度的AP结果相差不大，而高精度的AP@IoU=0.8结果相比GACHIPS要低18.2%。这些正样本点在l_{\triangle xy}^{reg}这一项loss上存在较大的偏差，并且无法继续优化。导致这种情况的原因是，多正样本在不同关键点位置的offset监督值不一致，越远离中心点offset监督数值越大，这些属于同一个目标的正样本点对整体的loss提供了大小迥异的梯度值。然而，梯度下降算法只能够将各个正样本点的各个通道回归到同等的数值水平，对于存在更大offset位置的正样本点，它们需要更高的权重，才能使得它们的预测值更接近真值。然而，去解决由于多正样本设计引入的这样的问题，收益并不如直接使用单正样本点设计，因为采用focal loss已经能够很好地解决目前场景下的前后景分类问题。2. 在单正样本设计中，为了同步分类和回归对模型梯度的贡献，采取了分类引导回归和分类回归绑定的策略。实验证明，原始的DIPS策略实际上是通过人工的方法定义了一组分类和回归的最佳正样本点，即二者都在监督框的中心点。这组点规避了样本分配操作，然而却存在有限的性能。基于分类引导回归的思路，GAHPS和GAHIPS两个实验表现出了更好的性能。二者基于当前轮次的训练分类结果，对回归的正样本点进行了位置定义。此时随着训练的进行，模型的回归正样本点将逐渐靠近分类正样本点。不同于DIPS，由于在训练过程中模型会对中心点以外的其它关键点进行回归，并且模型存在分类并不会完全收敛到目标中心点的理想情况（即中心点的分类分数并不一定会最高），因此在训练一定轮次之后，模型在测试阶段会基于当前的结果推理出当前的最佳预测bounding box，其存在最高的分类分数和最小的回归偏差，因此性能出现了增长。另外，从GAHIPS实验结果中发现，当回归预测结果参与样本分配时，模型的高精度性能进一步提升。这表明旋转IoU通过样本分配的方式参与了模型训练的引导，从而更进一步地提升了模型的性能。基于此，我们认为应该抛弃中心点假设，以完全自由的方式进行训练中正负样本的分配搜索。GACHIPS以Gaussian区域内每一个候选点解码的heatmap和IoU加和为cost，寻找到最佳的候选点作为当前目标的正样本点，来引导模型训练，最大化了分类和回归的一致性。实验表明，该方法使得模型总体性能进一步提升，并且在高精度性能上也实现了1.9%的提升。
% 要重新写-2024-01-17-yyl
% 已重写-2024-01-17-yyl
% 1. 试验表明多正样本策略会降低模型的高精度表现。
Experiments indicate that a \textbf{multi-positive sample} strategy can reduce a model's high-precision performance.
% 2. 原因是对同一个真值进行的多个多正样本loss计算会引入样本间的竞争。
The reason is that calculating multiple losses for the same ground truth object introduces competition between samples.
% 3. 具体而言，目前设计的loss中有与样本位置相关的delta x delta y。他们在不同的位置有不同的监督值，离中心更远的样本点处有更大监督值。
Specifically, the current loss design includes $\triangle{x}$ and $\triangle{y}$, which are related to the sample's position. These supervision values vary by location, with points further from the center having larger values.
% 4. 这意味着相同的真值目标的不同正样本点的回归终点是不一致的。这导致了模型在极值附近无法很好地收敛。
This means that different positive samples for the same ground truth object have inconsistent regression endpoints, preventing the model from converging well near the extremes.
% 5. 在传统的图像检测中，多正样本点策略很流行。但它们的出发点多为在互相遮挡的条件下，提升目标的检出率。然而，这一点并不能适用于BEV下的自动驾驶目标感知，因为目标在BEV下不存在相互遮挡。
In traditional image detection, multi-positive sample assignment strategies are popular for improving detection rates under overlap conditions. However, this approach is not suitable for BEV perception in autonomous driving, as targets do not overlap each other in BEV.

% 要重新写-2024-01-17-yyl
% 已经重写-2024-01-17-yyl
% 1. 然而，单正样本点的假设则很好地避免了这种竞争，为模型创造了单一的收敛终点。
However, the assumption of a single positive sample point effectively avoids this competition, creating a singular convergence point for the model.
% 2. 但是，单正样本点的前后景分类和框回归如果存在不一致性，也会限制模型的极限性能。
Yet, inconsistencies between foreground-background classification and bounding box regression with a single positive sample point can limit the model's ultimate performance.
% 3. 具体而言，在采用DIPS策略时，模型采用了固定的目标中心点来计算框回归损失，而推理的前后景分类最优点并不一定在目标中心。因此会造成模型精度的损失。这样的设计存在典型的分类回归不一致性问题。
Specifically, when employing the \textbf{DIPS} strategy, the model uses a fixed target center point to calculate box regression loss, but the optimal point for foreground-background classification during inference may not be at the target center, leading to a loss in model accuracy. This design has a typical inconsistency issue between classification and regression.
% 4. 因此，GAHPS策略希望找到一种方式，能够满足在训练结束后，推理的最佳前景点也同样经过了最多的框回归训练。因此，在每次训练时，前向推理的heatmap极值点被挑选为框回归的锚点。这样的策略的确带来了模型性能的提升。相比于DIPS，它的IoU @0.8高精度性能提升了xx%。
Therefore, the \textbf{GAHPS} strategy aims to find a way to ensure that the best foreground point during inference is also the one that has undergone the most box regression training by the end of training. Hence, during each training session, the peak points of the forward inference heatmap are selected as the anchor points for box regression. This strategy indeed brought an improvement in model performance. Compared to DIPS, its high-precision IoU 0.8 performance increased by 5\%.

% 5. 观察到上述的性能提升现象之后，我们猜测更好的cost设计将会进一步提升模型的精度。因此，我们将原本的heat map 替换成heatmap +IoU 。希望能够进一步缩小不一致性。试验表明这样的理解是正确的，模型确实在高精度性能上实现了进一步提升。相比GAHPS，GAHIPS在IoU @0.8上，提升了xx%。
After observing the aforementioned performance improvements, we hypothesized that a better cost design would further enhance model accuracy. Thus, we replaced the original heatmap with a heatmap plus IoU, hoping to further reduce inconsistencies. Experiments show that this understanding is correct, and the model indeed achieved further improvements in high-precision performance. Compared to GAHPS, \textbf{GAHIPS} improved by 2.5\% at IoU 0.8.
% 6. 更进一步地，我们尝试让这个被选择正样本同时控制前后景分类和框回归loss，我们提出了GACHIPS。这个修改意味着我们理论上实现了二者一致性的最大化。具体而言，在训练阶段，模型利用推理得到的最佳锚点进行loss计算。而在推理时，又能够通过该最佳锚点解码出最佳的检测框。这样的设计实现了前后景分类和框回归的统一性。也保证了训练和推理阶段的一致性。试验表明，该方案实现了最佳的高精度性能，在IoU@0.8上达到了93.90%。
Furthermore, we attempted to have the selected positive sample control both the foreground-background classification and the bounding box regression loss, proposing \textbf{GACHIPS}. This modification means that we have theoretically maximized the consistency between above two calculations. Specifically, during the training phase, the model uses the best anchor point obtained from inference for loss calculation. Then, during inference, it decodes the best detection box through this optimal anchor point. This design achieves unity between foreground-background classification and bounding box regression, ensuring consistency between the training and inference stages. Experiments show that this approach achieved the best high-precision performance, reaching 93.9\% at IoU 0.8.

%\textcolor{blue}{(From the experiment results, the conclusion is found that enhancing the relation between classification and regression can result a better performance of high precision. Furthermore, the method to bind the classification and regression together in loss procedure presents best effectiveness. However, unlike object detection on images, the multi-positive label assignment performs unsatisfactory. The reason is that, adding the number of positive sample actually enlarge the inconsistency between classification and regression. Because the loss is designed as a sum product of the classification and regression, and regression part consists of box length, box width, box angle, etc. (Since the model will tend to gradient-descend the loss value calculated by all positive sample to the same numerical level. Assuming that the prediction of all channels including classification and regression randomly initialized from a certain same numerical level, then the anchor points who locates with larger offsets would have much more difficulty to regress to a ideal level comparing with those who locates in the centers of the boxes. However the classification results are only needed to regress to 1, so there are not such a big numerical gap between prediction and supervision in classification). Therefore, a better regression result may not necessarily wins in the NMS decoding stage. This leads to the phenomena in the experiment, that is, the low-precision AP results are not much different, while the high-precision AP result are much lower about 10\% compares to the DIPS.)}

\subsubsection{Ablation Study on Interactive Transformer-based BEV Fusion Approaches}

To illustrate the effectiveness of interactive transformer-based BEV fusion approaches, both the direct transformer and interactive transformer are applied in fusion module with DIPS label assignment, respectively. Table \ref{ablation_study_ralibev} presents the results of transformers as well. In the \textbf{Direct Transformer for BEV Fusion}, the performance reached the same level as GACHIPS with only slightly lower performance, around 1.6\% in high precision IoU of 0.8. While the \textbf{DQMITBF} exhibits stronger performance, all three indicators exceed GACHIPS in an all-around way. This brings the AP result of IoU 0.5, 0.65, and 0.8 to 98.8\%, 97.7\%, and 94.8\%, respectively.

To demonstrate the effectiveness of DQMITBF in a more comprehensible manner, the feature maps of two selected two frames in the interactive transformer fusion module are visualized as depicted in Fig. \ref{Interactive transformer feature map visualization}. In Fig. \ref{Interactive transformer feature map visualization}, The top two rows are corresponding to Frame 2799 and the bottom two rows are corresponding to Frame 3299, which are selected randomly. Each frame includes selections from clear and foggy weather conditions for analysis. For each column, the feature maps $\bm{X}_Q, \bm{X}_R, \bm{X}_L, \bm{X}_Q\bm{X}_R^T$, and $\bm{X}_Q\bm{X}_L^T$ are illustrated for visual comprehension. Furthermore, all feature maps are normalized to the range $[0, 1]$ for a consistent representation.

Utilizing a unified BEV query $\bm{X}_Q$, the input radar feature map $\bm{X}_R$ and LiDAR feature map $\bm{X}_L$ are weighted to $\bm{X}_Q\bm{X}_R^T$ and $\bm{X}_Q\bm{X}_L^T$ respectively, subsequently producing a fused feature for additional processing. It can be observed that $\bm{X}_Q$ manages radar and LiDAR features differently under varying weather conditions. In clear weather, $\bm{X}_Q\bm{X}_L^T$ is noticeably larger in numerical value than $\bm{X}_Q\bm{X}_R^T$. However, in foggy conditions, the input $\bm{X}_L$ is significantly impacted by noise, resulting in $\bm{X}_Q\bm{X}_R^T$ being larger than $\bm{X}_Q\bm{X}_L^T$. From these observations, we can draw the following conclusions:

\begin{itemize}
    \item The approach utilizes an interactive transformer to aggregate data from LiDAR and radar branches in a unified manner;

    \item The BEV query in the interactive transformer can focus on more information-rich signals under different weather conditions.
\end{itemize}

Additionally, as depicted in Fig. \ref{Interactive transformer feature map visualization}, the input $\bm{X}_L$ and $\bm{X}_R$ appear visually similar to the raw source data, and $\bm{X}_Q$ does not seem to focus on a specific area determined by the presence of an object. This is because the interactive transformer fusion module operates in the encoder process of the entire model at an early stage,  where the data feature resides in a raw energy distribution space rather than a box location feature space.

The result of the experiment shows that a reasonable fusion logic would significantly enhance the performance of the proposed RaLiBEV.
While direct transformer fusion among feature maps can enhance performance, the implementation of interactive transformer-based concepts can yield superior results. This is largely due to the innovative querying fusion method, which introduces symmetry and unifying properties into the model.

To also highlight the benefits of multi-modality fusion, Table \ref{radar_lidar_vs_lidar_only} presents a comparison of performance metrics between the LiDAR-only approach and the fusion of radar and LiDAR. It is evident from the comparison that neither RaLiBEV with LiDAR only nor PointPillars can match the performance levels achieved by the fusion of radar and LiDAR. This performance enhancement is particularly noticeable in foggy testing scenarios. When the model is trained on a ``Clear+Foggy" dataset, RaLiBEV, in the radar and LiDAR fusion modality (denoted as ``R+L"), exhibits exceptionally high performance, especially under the stringent requirements of a 0.8 IoU in ``Foggy" test scenarios.

\subsubsection{Performance Comparison with State-of-the-Art Methods on Different Weather Conditions}

\begin{figure*} [htb]
    \center
    \includegraphics[scale=1.1] {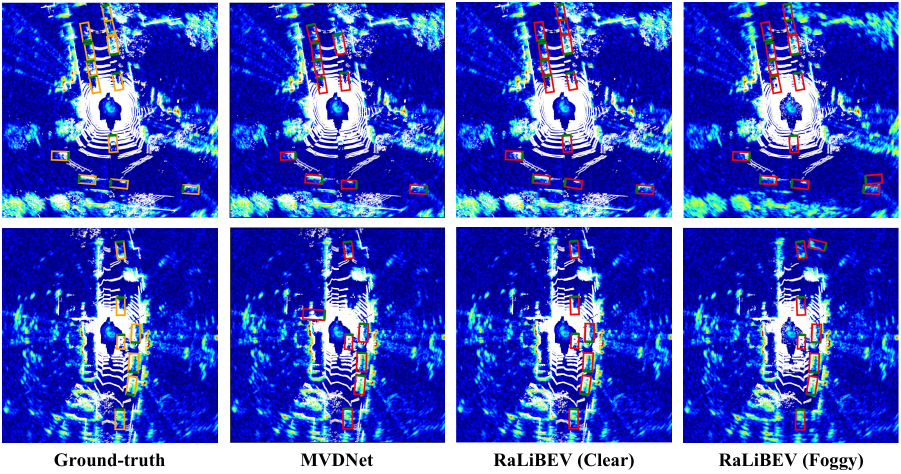}
    \caption{Visualization of detection results of MVDNet and RaLiBEV.}
    \label{RaLiBEV prediction result compare with MVDNet}
\end{figure*}

Following MVDNet \cite{radar_LiDAR_fusion_object_detection_MVDNet}, model performance in different weather is tested in this paper as described in Table \ref{sota_comparison}. The simulated foggy LiDAR data is generated by the code provided by MVDNet. This experiment is set as training using ``Clear-only" and ``Clear+Foggy" LiDAR and radar data, then testing under both ``Clear" and ``Foggy" data. Two peak performance models are chosen to prove their capability under this condition. One is RaLiBEV with GACHIPS, the other is RaLiBEV with DQMITBF. The performance of GACHIPS training under ``Clear-only" data and testing under the ``Foggy" scenario is better than DQMITBF's. The performance of the horizontal comparison model after data enhancement is about 10\%. GACHIPS increases from 84.6\%, 82.4\%, and 73.8\% to 97.8\%, 96.7\%, and 93.7\% under the ``Foggy" test. This result exceeds the ST-MVDNet \cite{radar_LiDAR_fusion_object_detection_ST-MVDNet} over 20\% at ``Clear+Foggy" training and``Foggy" test scenario. Compared to the newest work Bi-LRFusion \cite{radar_LiDAR_fusion_object_detection_Bi-LRFusion}, RaLiBEV outperforms a large margin of 20.7\% in IoU 0.8 requirement. Meanwhile, according to the results of the state-of-art work ST-MVDNet++ \cite{radar_LiDAR_fusion_object_detection_ST-MVDNet++}, RaLiBEV has made progress of 4.4\%, 7.8\% and 19.0\% in IoU 0.5, 0.65 and 0.8 under ``Clear+Foggy" training, ``Foggy" test. This result represents that RaLiBEV is able to perform better accuracy in adverse weather such as fog. The fusion logic and label assignment strategy help the detector achieves higher precision.

Fig. \ref{RaLiBEV prediction result compare with MVDNet} shows the ground-truth, the prediction from MVDNet in clear weather, the prediction from RaLiBEV in clear weather and foggy weather at the $3332^{th}$ and $4103^{th}$ frame of test dataset. The ground-truth bounding boxes are decorated with yellow color, and the predicted boxes by each model are shown with red color. All boxes use green lines to indicate the heading of the object. It is easy to find out that the prediction of MVDNet have 3 miss detection and 4 predicted boxes with reversed heading direction in the top row figure, and 1 false alarm and 2 reversed predicted boxes in the bottom row figure, though the weather is clear. In contrast, The RaLiBEV shows perfect detection results in clear weather, and only gives one false detection and no reversed heading prediction in both time frame under foggy weather.

% \iffalse
% \begin{table*}[ht]
% \caption{Model Performance Between Transformer Block and Label Assignment Strategy in Several Aspects.} % \vspace{-15pt}}
% \begin{center}{
% \scalebox{1.0}{
% \begin{tabular}{|c|c|c|c|c|}
% \hline
% & GFLOPs & Training Time & Inference Time & \#Params \\ \hline
% Direct Index-based Positive Sample Assignment (DIPS)                                   &  18.451  &               & 0.674          &  60.822M  \\ \hline
% Gaussian Area-based Posterior Heatmap and IoU Cost Positive Sample Assignment (GACHIPS) & 18.451  &               & 0.609          &   60.822M       \\ \hline
% Direct Transformer for BEV Fusion                                           &   18.340    &               & 0.670          &  60.821M        \\ \hline
% Dense Query Map-based Interactive Transformer for BEV Fusion(DQMITBF)           &  18.811   &               & 0.621          &   60.826M       \\ \hline
% \end{tabular}
% }
% }\end{center}
% \end{table*}
% \fi

%Tao: 这里感觉有点短了。
\section{Conclusion}\label{sec conclusion}
% highlight最重要的影响，foggy 下的 high accuracy。最希望让他记住什么。销售技巧。
In this paper, we introduce RaLiBEV, an innovative method for fusing LiDAR and radar data. According to the experiment results, this method effectively showcases the significant role that radar can play in fusing autonomous driving perception technology, especially in adverse weather.
% 创新点写在这里，实验结果写在这里，
% 在solve xxx by propose xxx
To enhance the performance of the model, especially for high-precision detection, novel label assignment and transformer-based fusion strategies are proposed. Specifically, the 
%
%Gaussian Area-based Consistent Heatmap and IoU Cost Positive Sample assignment 
%
GACHIPS strategy is proposed to dynamically adjust the classification and regression positive sample position, to solve the inconsistency problem in the loss procedure. Meanwhile, the 
%
%Dense Query Map-based Interactive Transformer for BEV Fusion 
%
DQMITBF is proposed to aggregate weighted features from different 
sensor data. Thanks to the interaction of the dense learnable query with radar and LiDAR in the symmetrical manner, RaLiBEV can swtich the focus on different sensor modalities adaptively, according to change of the perception environment and weather conditions.
It's also noteworthy that by incorporating the GACHIPS label assignment into a simple fusion architecture, the model can attain performance on par with the more intricate transformer fusion.
The experiments with ORR dataset show that RaLiBEV exceeds all existing state-of-the-art approaches by a large margin.

%\input{Appendix}

% or


% \appendices
% \section{Network Architecture}


% \section{}

% \appendix[Network Architecture]

%\section*{Acknowledgment}


\ifCLASSOPTIONcaptionsoff
  \newpage
\fi

%%%%% ----- references section

%\bibliographystyle{IEEEtran}
%\bibliography{reference}

%\iffalse
%\newpage

%\input{Bio}

% \fi

%\iffalse
%\newpage

%\fi

\end{document}